\documentclass{article}

\usepackage[preprint]{corl_2026} 
\usepackage{tabularx}
\usepackage{array}
\usepackage{booktabs}
\usepackage{tabularx}
\usepackage{multirow}
\usepackage{graphicx}
\usepackage{amsmath}
\usepackage{amsfonts}
\usepackage{enumitem}
\usepackage{url}
\usepackage{microtype}
\usepackage{xspace}
\usepackage{bm}
\usepackage{threeparttable}
\usepackage[table]{xcolor}
\usepackage{pifont}
\usepackage{wrapfig}

\usepackage{xcolor}
\usepackage{hyperref}
\usepackage{fancyhdr}
\definecolor{hiphipurple}{RGB}{178,49,144}
\newcommand{\ours}{HiPHI\xspace}

\fancypagestyle{noitomfirst}{%
  \fancyhf{}%
  \fancyhead[L]{\includegraphics[width=0.52\textwidth]{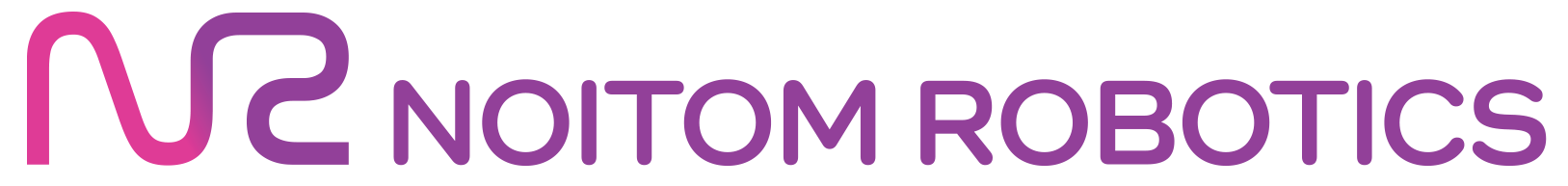}}%
  \fancyhead[R]{\raisebox{0.35\baselineskip}{\large\itshape August 2026}}%
}
\AtBeginDocument{%
  \addtolength{\topmargin}{-17pt}%
  \setlength{\headheight}{29pt}%
}

\title{\ours: A Large-Scale Benchmark for High-Precision Human Motion and Object-Interaction}

\author{
  \textbf{
    Jiahao Ji$^{1,2,*}$,
    Ji Ma$^{1,5,*}$,
    Runhan Zhang$^{1,3,*}$,
    Runyi Yu$^{1,4}$,
    Wenjia Wang$^{1,5}$,
    Weiheng Chi$^{1}$,
  } \\
  \textbf{
    Qianqian Peng$^{1}$,
    Weichao Yan$^{1}$,
    Yongfei Gu$^{1}$,
    Ye Tian$^{1}$, 
    Ting Wu$^{1}$,
    Longwei Li$^{1}$,
  } \\
  \textbf{
    Chun Yuan$^{3}$,
    Ruoli Dai$^{1,\dagger}$,
    Lei Han$^{1,\dagger}$
  }
  \\[4pt]
  $^{1}$Noitom Robotics\\
  $^{2}$National University of Singapore\\
  $^{3}$SIGS, Tsinghua University\\
  $^{4}$The Hong Kong University of Science and Technology\\
  $^{5}$The University of Hong Kong
  \\[4pt]
  $^{*}$Equal contribution. \\
  $^{\dagger}$Corresponding authors. \\
  \texttt{[tristan@noitomrobotics.com], [lhan@noitomrobotics.com]}
}

\begin{document}
\maketitle
\thispagestyle{noitomfirst}
\vspace{-3em}

{\centering
\small
Project page:
\href{https://noitom-robotics.github.io/hiphi}{%
  \textcolor{hiphipurple}{https://noitom-robotics.github.io/hiphi}%
}\par
}

\vspace{-0.6em}

\begin{center}
    \includegraphics[width=0.8\linewidth]{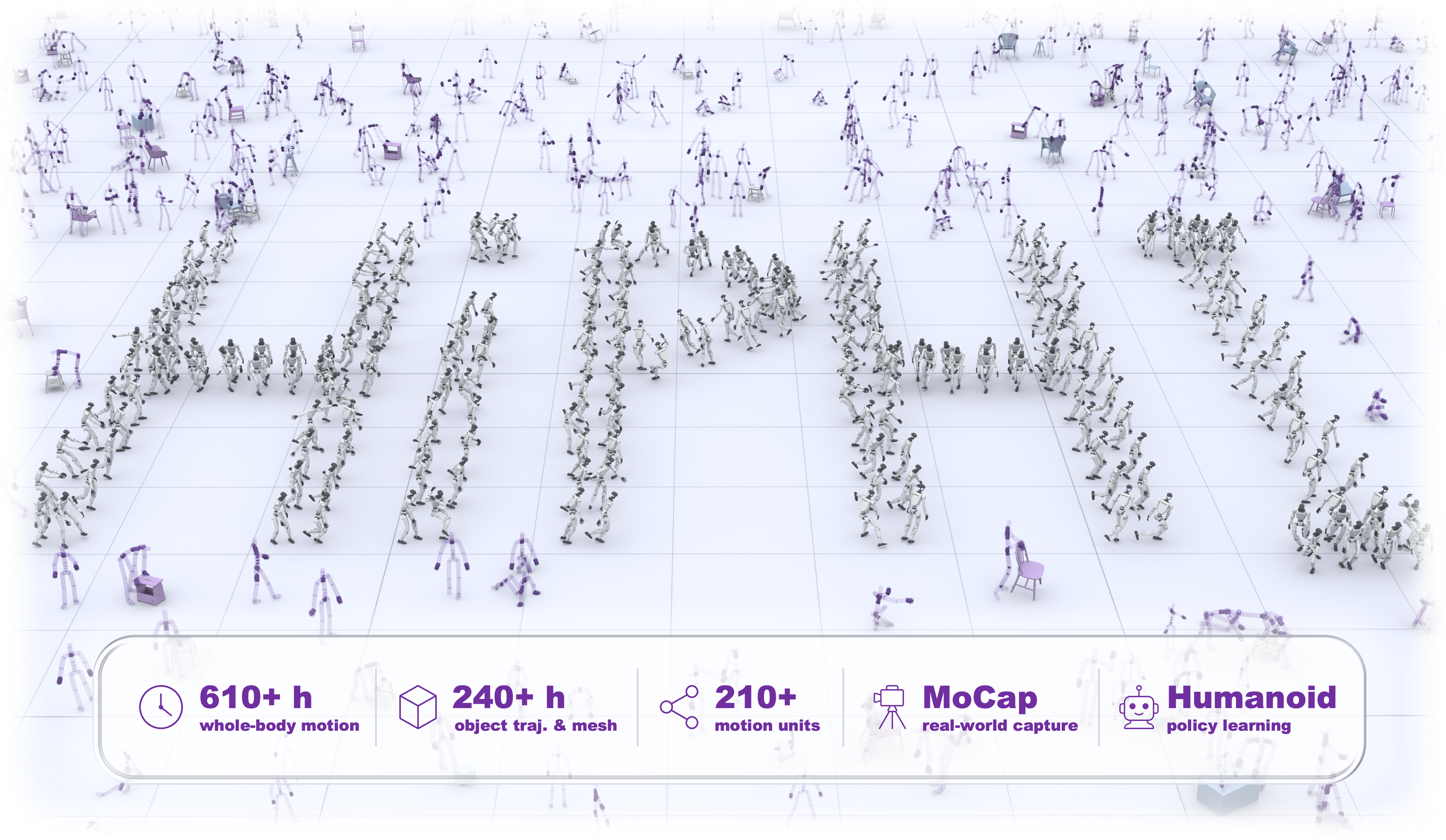}
    \vspace{-1.6em}
    \refstepcounter{figure}
    \label{fig:teaser}

    \begin{minipage}{0.8\linewidth}
    \small
    \vspace{0.15in}
    Figure~\thefigure: \textbf{Overview of \protect\ours{}.}
    A large-scale, high-fidelity dataset for humanoid learning.

    \end{minipage}
\end{center}

\begin{abstract}
Humanoid intelligence requires learning over an extremely diverse space of whole-body motions and physically grounded interactions. However, existing embodied datasets remain fundamentally limited: internet-scale video data lack precise physical states and interaction grounding, while laboratory motion datasets provide high fidelity but only narrow behavioral coverage. This mismatch creates a critical bottleneck for scalable humanoid policy learning.
We present \ours, a 600+ hour scale high-fidelity whole-body human motion dataset designed to systematically maximize coverage of the human motion and interaction manifold.
\ours is theoretically guided by FrameNet, a linguistic framework organizing human primitives.
Created using an optical motion capture pipeline, \ours provides sub-millimeter spatial marker tracking accuracy for full-body human motion and mesh-level object trajectories.
We further introduce a benchmark suite evaluating motion-space diversity, interaction grounding, object consistency, and physical AI applications. Our analyses demonstrate that \ours significantly expands motion coverage compared to existing motion datasets while maintaining high-fidelity interaction quality, and establishes a scalable data foundation for training, evaluating, and generalizing humanoid policies in real-world embodied tasks,  where similar extensions are also applicable to motion prior models in computer graphics.
\end{abstract}


\section{Introduction}
\label{sec}

Humanoid learning requires reference motion that is both broad in coverage and physically faithful. Robot policies must acquire balance, posture transitions, limb coordination, contact, and load-dependent strategies. However, most existing data sources are not designed to capture high-precision human behavior for humanoid learning. Robot demonstrations and teleoperation data are limited by scale and cost, and are tied to specific embodiments~\citep{o2024open,chi2024universal}. Human videos and egocentric data provide broad behavioral diversity at a large scale, but are limited to visual observations and therefore need to rely on reconstruction or proxy signals rather than directly measured physical states~\citep{grauman2022ego4d}. MoCap and HOI datasets provide accurate motion and sometimes object state, but are primarily designed for motion synthesis, language grounding, reconstruction, or interaction understanding~\citep{mahmood2019amass,guo2022generating,bhatnagar2022behave,lu2025humoto}. Humanoid policy learning, therefore, calls for data that are broad, precise, and grounded in object-constrained whole-body control.

We formulate human data collection for humanoid learning as a problem of motion-space design. Rather than manually curating behavioral scripts, \ours uses FrameNet~\citep{baker1998berkeley,ruppenhofer2016framenet} as a semantic scaffold, where frames represent event types and lexical units (LUs) identify the word senses that evoke them.
We select frames and LUs most relevant to embodied intelligence, and use each LU under a specific frame as a seed for performer-facing capture scripts. We then expand each seed along multiple dimensions, including direction, speed, amplitude, body posture, body-part involvement, and object/contact conditions, producing a diverse set of repeatable motion instances for systematic capture. This turns handcrafted script authoring into a structured and scalable process for expanding motion-space coverage.

Using this pipeline, we present \ours, a 600-hour-scale, high-precision human motion and object-interaction benchmark, all captured from 132 performers using a large optical motion capture (MoCap) system with sub-millimeter spatial tracking accuracy. Specifically, \ours combines 371.8 hours of diverse short-horizon whole-body motion and 245.7 hours of real-object interaction data, spanning 214 Frame–LU motion units across 22 frames. The interaction subset covers 40 distinct real-world objects from 12 categories. Each clip is indexed by a Frame–LU label and paired with a natural-language description. For human-object interaction sequences, synchronized object trajectories and meshes are captured together with human motion, making object state an integral part of the motion record. Capturing real objects is essential, as load, friction, inertia, resistance, and contact substantially shape control strategies in ways that pantomime motion cannot reflect.

We evaluate \ours in terms of motion-space coverage, data quality, interaction consistency, and humanoid policy learning performance. Motion coverage is measured using label-free kinematic embeddings; data quality is assessed through smoothness, ground contact, and interaction geometry; and downstream applications are tested through humanoid motion tracking and real-world deployment. Together, these analyses show that \ours is diverse, physically stable, interaction-grounded, and usable as executable reference data in humanoid learning, substantially surpassing existing datasets.

Our contributions are threefold: (i) a FrameNet-guided motion-space construction pipeline for systematic and scalable coverage of whole-body motion and interaction; (ii) a 617.5-hour, high-precision dataset including 245.7 hours of human–object interaction, synchronized object trajectories and meshes, Frame–LU indexing, and natural-language descriptions; and (iii) a comprehensive evaluation protocol for humanoid robot learning, including motion coverage, data quality, interaction consistency, motion tracking, and real-robot deployment.

\section{Related Work}
\label{sec:related}

\textbf{Robot Teleoperation and Egocentric Vision Data.}
Embodied intelligence increasingly demands large and diverse data sources. One major category is real-robot data, which avoids the embodiment gap, including teleoperated robot data~\citep{o2024open,khazatsky2024droid,walke2023bridgedata}, manipulation benchmarks~\citep{james2020rlbench,li2023behavior,mees2022calvin,liu2023libero}, and low-cost interfaces such as UMI~\citep{chi2024universal}. These data sources avoid gaps in robot embodiment or end-effector configuration, but are often limited by collection cost, data scale, behavioral diversity, and dependence on specific robots, hardware, sensors, and viewpoints. Recently, egocentric visual data have become increasingly popular, as they provide life-scale observations of long-tail human behaviors in the real world~\citep{grauman2022ego4d,kareer2025egomimic,hoque2025egodex,li2025h2r,zhu2026emma}. However, such data are typically limited to visual modalities, making it difficult to recover precise human motion and physical interaction states.

\textbf{Human Motion Capture Data.}
Motion capture data can accurately record human motion, with widely used datasets including AMASS~\citep{mahmood2019amass}, LAFAN1~\citep{harvey2020robust}, BABEL/KIT/HumanML3D~\citep{punnakkal2021babel,plappert2016kit,guo2022generating}, Motion-X/Motion-X++~\citep{lin2023motionx,zhang2025motion}, and the recent BONES-SEED~\citep{bonesseed2026}. These datasets were originally designed mainly for motion synthesis, annotation, generation, or animation, but have recently become widely used in humanoid learning. However, most were not systematically designed for embodied intelligence: few simultaneously provide large scale, high-precision motion, systematic motion-space coverage, and synchronized object states. In contrast, \ours is a large-scale dataset specifically designed for embodied intelligence and humanoid learning. It uses the well-established FrameNet theory to structure data collection and directly evaluates motion-space coverage.

\textbf{Human-Object Interaction.}
Similarly, many HOI datasets have been introduced for motion synthesis, animation, and interaction understanding~\citep{taheri2020grab,bhatnagar2022behave,huang2024intercap,liu2022hoi4d,li2023omomo,lv2024himo,lu2025humoto,xu2025interact}. However, these datasets are typically small in scale, often only a few hours long, and their contact-level accuracy and interaction diversity remain limited. In contrast, our dataset contains 245.7 hours of HOI data and explicitly emphasizes the physical accuracy of contact-rich interactions. This design supports efficient humanoid learning based on motion imitation methods~\citep{peng2018deepmimic,peng2021amp,peng2022ase,wagener2022mocapact,luo2023perpetual,zhu2023neural,han2024lifelike,liao2025beyondmimic,luo2025sonic,yang2025omniretarget,protomotions2025}. 

\section{FrameNet-Guided Motion-Space Construction}

\label{sec:dataset_construction}
\begin{figure*}[t]
    \centering
    \includegraphics[width=0.95\textwidth]{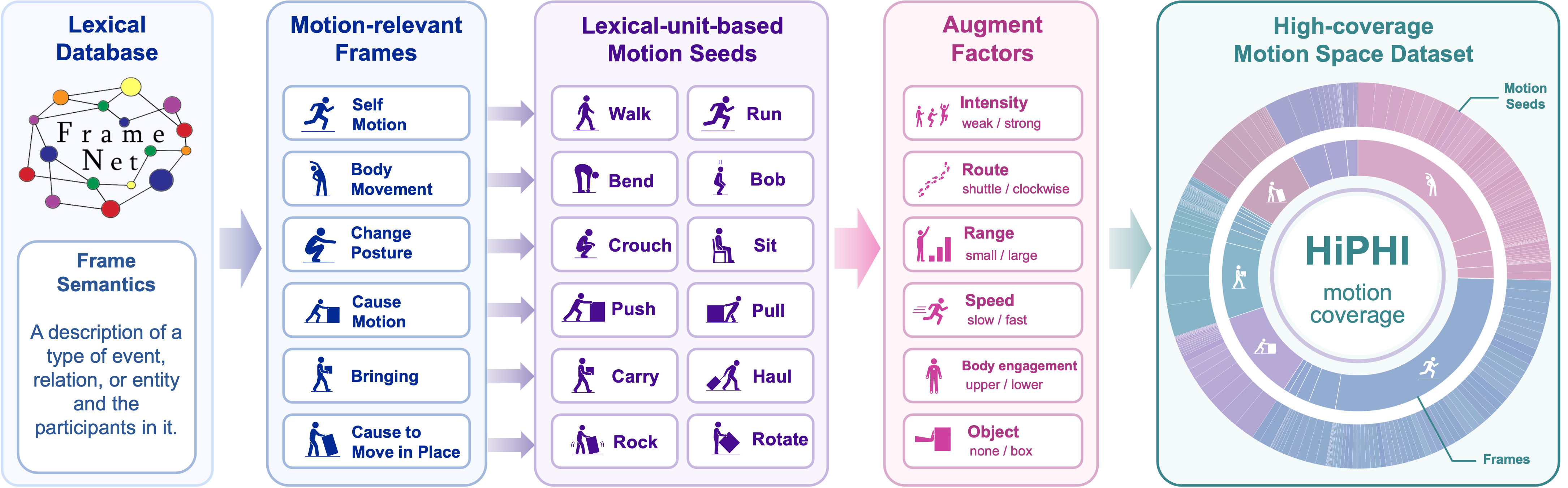}
    \caption{
    \textbf{From FrameNet to \ours data construction.} Motion-relevant frames and lexical units define motion seeds, which are expanded through controlled factors such as speed, direction, amplitude, body-part involvement, and object/contact conditions.
    }
    \label{fig:framenet_to_realizations}
\end{figure*}

\subsection{Motion Coverage}
\label{sec:motion_coverage}

The most suitable data for humanoid robot learning is not necessarily the full distribution of motions that appear in everyday human life. For example, we do not expect robots to over-learn behaviors such as sleeping, eating, or watching television, which robots do not need to perform. Instead, the data most relevant to humanoid learning should cover control-centric motion patterns, including locomotion, posture transitions, limb coordination, dynamic balance, and responses to external constraints. The goal is therefore not to reproduce the full catalog of human activities, but to cover the motion space that matters for humanoid learning.

On the other hand, the human behaviors or tasks in existing motion datasets are often derived from manually designed scripts. These scripts usually describe semantically meaningful daily actions, which makes the resulting motion clips natural and interpretable. However, because the tasks are manually designed at the semantic level, they can lead to substantial overlap in motion space. For example, ``raising a hand to wipe sweat from the forehead'' and ``raising a hand to block sunlight'' are two different scripts at the task level, but may produce nearly identical motions. Counting them as two different tasks does not give us two genuinely different motions. The reverse is also true: one simple action such as walking can produce many different whole-body motions when its direction, speed, stride length, turning pattern, or posture changes. A growing list of manually designed scripts therefore provides no clear way to tell which parts of the motion space have already been covered and which are still missing. For humanoid learning, we instead aim to enable robots to span the full reachable human motion space, so that learned policies can better adapt to diverse downstream tasks. This requires a structured set of motion units that can be systematically enumerated and expanded, rather than an open-ended list of task descriptions.

Finally, existing datasets often contain many human-object interaction motions without capturing the corresponding object states, leading to, for example, ``a person sitting in the air''. In this example, the body trajectory is recorded, but the chair that provides the support surface and contact relation is missing, leaving the physical interaction incomplete. Data for object-constrained humanoid learning should therefore capture the relevant object states together with the human motion. For instance, pushing, carrying, dragging, or leaning can substantially change posture, foot placement, and center-of-mass motion through the object's geometry, friction, load, and trajectory. These factors are essential for learning physically grounded humanoid behaviors. \ours records these objects together with the human body, preserving the complete interaction as physically executable motion reference.

\subsection{A FrameNet-Guided Collection Pipeline}
\label{sec:framenet_enumeration}

To systematically address the problem of motion-space coverage, we use FrameNet as a semantic foundation for motion construction. FrameNet organizes word meanings according to the events they describe. A frame represents a type of event, while a lexical unit (LU) represents a word used in one particular meaning within that frame. For example, \textit{walk}, \textit{jog}, and \textit{run} under the \textit{Self\_motion} frame describe related forms of self-propelled movement. This distinction is important because the same word may describe completely different events: ``run across the field'' refers to body motion, whereas ``run a company'' does not. A Frame–LU pair identifies the first meaning precisely instead of treating every use of the term ``run'' as the same motion.

This follows the same organizing idea as ImageNet~\cite{deng2009imagenet}, where WordNet~\cite{miller-1994-wordnet} provided a structured set of visual concepts to collect: FrameNet provides \ours with a structured set of motion meanings to collect. The goal is to avoid relying on manually designed task scripts and instead enumerate the motion space from a theoretically grounded and well-structured system. The action lexical units in FrameNet naturally serve this purpose, making them suitable seeds for generating a broad motion space. Figure~\ref{fig:framenet_to_realizations} shows this complete pipeline from the FrameNet scaffold to the captured \ours motion space.

We select frames and lexical units that are most relevant to humanoid robots, including those related to body motion, posture change, directional movement, body-part motion, object actuation, and human-object interaction. Starting from these units, we further expand them along dimensions such as path, direction, speed, rhythm, amplitude, body posture, body-part involvement, support relation, and object/contact conditions. For example, a walking seed can be expanded with different routes, directions, speeds, stride lengths, turning patterns, and postures. A pushing seed can vary the object, load, contact point, pushing direction, and object trajectory. These variations change the actual human or object motion, rather than merely changing the story attached to it. These expanded scripts are then provided to motion-capture performers for execution. In this way, each Frame–LU pair maps to a family of capturable motions rather than a single clip.

\begin{figure}[t]
    \centering
    \includegraphics[width=0.7\linewidth]{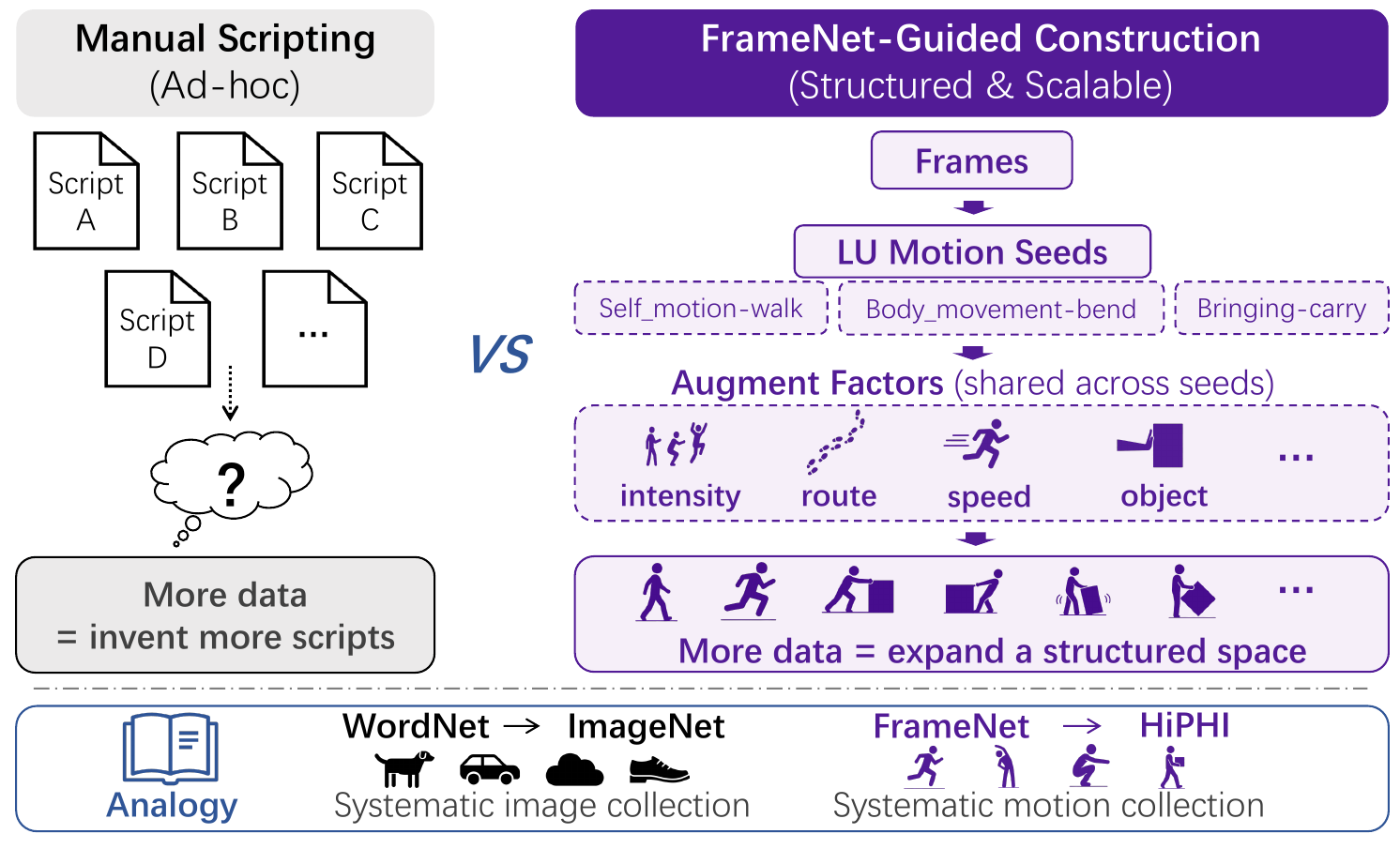}
    \caption{
    \textbf{Manual scripting versus FrameNet-guided construction.}
    Manual collection grows by adding scripts one by one, whereas \ours organizes collection around Frame-LU motion seeds and expands them through shared factors.
    This turns dataset growth into a structured and scalable expansion of the motion space.
    }
    \label{fig:framenet_comparison}
\end{figure}

Figure~\ref{fig:framenet_comparison} makes the difference from manual script design explicit. In a conventional workflow, scripts are typically written case by case, guided mainly by the designers' experience and intuition, without an explicit taxonomy or coverage criterion. Dataset growth therefore becomes an ad hoc, trial-and-error process: there is no principled way to determine which motions are still missing, whether a new script genuinely expands the motion space, or whether it merely wraps an already collected motion in a different story. \ours instead starts from Frame-LU motion seeds and expands them through a shared set of factors, such as intensity, route, speed, and object conditions. This replaces the blind accumulation of disconnected scripts with a systematic and traceable expansion of motion units and their physical variations. To our knowledge, \ours is the first MoCap dataset for robot learning to prospectively adopt such a linguistic scaffold for data construction.
\section{Dataset Composition and Statistics}
\label{sec:dataset_statistics}

Following this construction procedure, \ours is created as a LU-indexed motion dataset containing body-only and object-interaction sequences. This section summarizes its scale, relationship to existing datasets, Frame-LU composition, and object-interaction subset. The file organization and metadata are detailed in Appendix~\ref{app:dataset_composition_metadata}.

\subsection{Data Characteristics}
\label{sec:data_charac}
\ours contains {617.5 hours} of high-fidelity optical motion capture data in BVH format, corresponding to approximately {200.1 million frames}. This is obtained by applying left-right mirroring to {308.7 hours} of original motion captured with a high-precision optical MoCap system, following the convention used by BONES-SEED~\citep{bonesseed2026}. Human motion is captured at {90 Hz} from 132 distinct performers, with detailed statistics in Appendix~\ref{app:dataset_composition_metadata}. For object-interaction sequences, the data include synchronized human motion, object trajectories, and object meshes, making the object state an integral part of the motion record.
Table~\ref{tab:dataset_comparison} compares \ours with existing representative motion and human-object interaction datasets. Existing data often emphasize either large-scale body motion without object state or smaller-scale object-centric interaction. In contrast, \ours combines large-scale high-precision optical MoCap, Frame-LU indexing, synchronized object state, and a benchmark protocol designed for humanoid robot learning.

\begin{table}[t]
\centering
\caption{\textbf{Comparison with representative human motion and interaction datasets.} \ours offers 617.5 hours of MoCap data, including 245.7 hours of object-state-aligned motion, that exceeds representative human motion and interaction datasets in scale and interaction coverage.}
\label{tab:dataset_comparison}
\small
\setlength{\tabcolsep}{3pt}
\renewcommand{\arraystretch}{1.08}
\begin{tabular*}{0.9\linewidth}{@{\extracolsep{\fill}}l r l l l l@{}}
\toprule
\textbf{Dataset} & \textbf{Hours} & \textbf{Capture} &
\textbf{Format} & \textbf{Motion index} & \textbf{Object motion} \\
\midrule
\textbf{\ours} & \textbf{617.5} & \textbf{MoCap} &
\textbf{BVH} & \textbf{Frame-LU} & \textbf{245.7~h} \\
AMASS & $>$40 & MoCap &
SMPL & -- & -- \\
BONES-SEED & 288.3 & MoCap &
SOMA/G1 & NL segments & -- \\
Motion-X++ (orig.)\textsuperscript{\dag} & 40.4 & Video &
SMPL-X & text + pose & -- \\
LAFAN1 & 4.6 & MoCap &
BVH & action themes & -- \\
GRAB & 3.8 & MoCap &
SMPL-X & intent labels & 3.8~h \\
OMOMO & 9.8 & MoCap &
SMPL-X & -- & 9.8~h \\
HIMO & 9.4 & MoCap &
SMPL-X & text seg. & 9.4~h \\
\bottomrule
\end{tabular*}
\vspace{0.5mm}
\begin{flushleft}
\footnotesize
\textsuperscript{\dag} Original Motion-X++ data only, where its integrated third-party sources are excluded. NL indicates Natural Language.
\end{flushleft}
\vspace{-1mm}
\end{table}

\subsection{Frame-LU Composition and Long-Tail Coverage}
\label{sec:frame_lu_composition}

\begin{figure*}[t]
    \centering
    \includegraphics[width=0.8\textwidth]{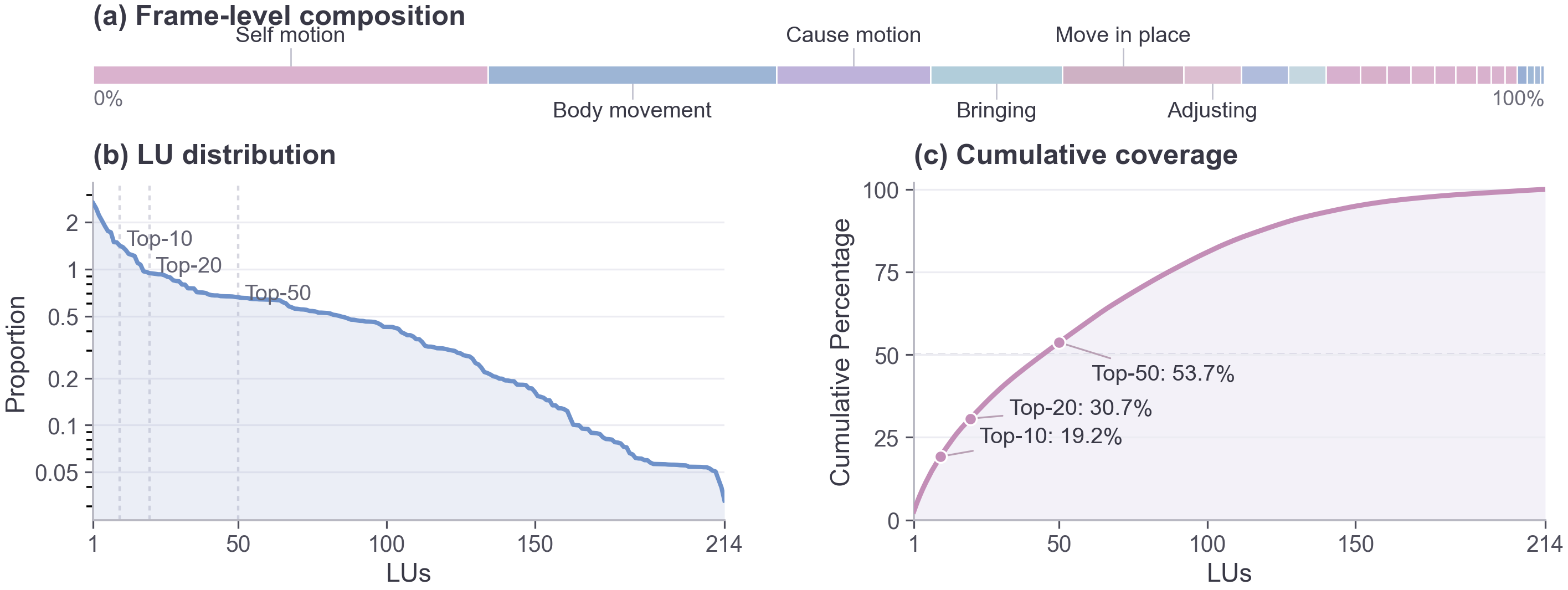}
    \caption{\textbf{Frame-LU composition and long-tail structure.}
    The figure shows duration by FrameNet frame, the Frame-LU duration distribution, and cumulative duration share, illustrating that \ours combines common motion units with a broad long tail of whole-body patterns.}
    \label{fig:frame_lu_long_tail}
\end{figure*}

Each clip in HiPHI is paired with a Frame-LU label and a natural-language description. The Frame-LU labels also serve as indices for retrieval, sampling, and analysis. The {214} Frame-LU labels across {22} FrameNet frames, placing each motion meaning within its event context rather than relying on ambiguous motion names. As shown in Figure~\ref{fig:frame_lu_long_tail}, the dataset spans both frequent motion units, such as locomotion, posture changes, and body-part movements, and a long tail of more dynamic, irregular, and constrained patterns. The top 50 Frame-LUs account for {53.7\%} of the released duration, leaving {46.3\%} distributed across the remaining long-tail motion units. Motion scripts generated from each LU are performed by approximately {24} different actors on average, and {154} Frame-LUs are performed by at least {10} actors, providing substantial performer variation within individual motion units.

\subsection{Body-Only and Object-Interaction Motions}
\label{sec:object_interaction_composition}

\ours includes {371.8 hours} of body-only motion data and {245.7 hours} of human-object interaction motion data. The body-only subset covers self-motion, posture changes, body-part movement, dynamic motion, and coupled whole-body coordination. The object-interaction subset records motions shaped by real geometry, contact, load, and object motion, such as sitting, leaning, supporting, pushing, pulling, and carrying. It contains {40} objects across {12} categories, ranging from furniture and containers to cleaning tools and sports equipment, with masses from {0.45} to {6.25} kg. The subset spans {90} Frame-LUs across {15} FrameNet frames. Together, \ours provide a large-scale reference base for physically grounded interaction analysis and object-constrained humanoid learning.
\section{Experiments}
\label{sec:experiments}

In this section, we create the HiPHI benchmark and evaluate \ours along two axes: dataset-level coverage and quality (Secs.~\ref{sec:motion_space_diversity},~\ref{sec:data_quality_precision}), and downstream utility through humanoid tracking in simulation and sim-to-real deployment (Secs.~\ref{sec:humanoid tracking}, ~\ref{sec:real world}).

\begin{figure*}[t]
    \centering
    \includegraphics[width=0.95\textwidth]{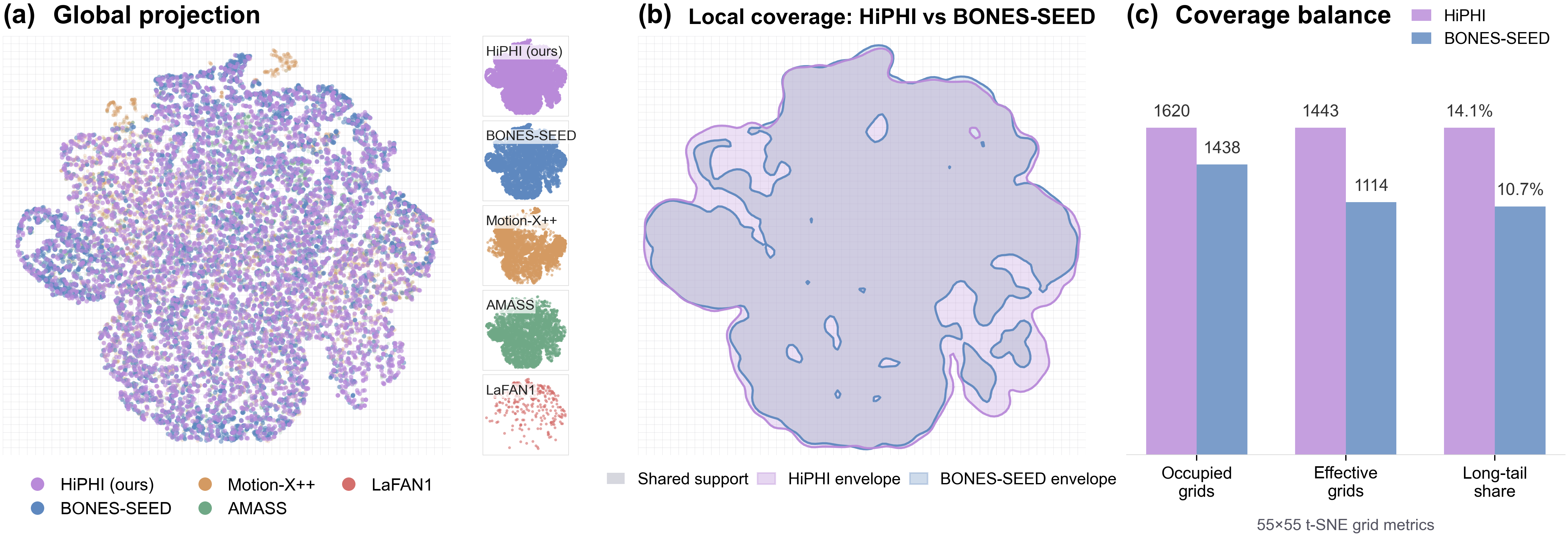}
    \caption{\textbf{Full kinematic motion-space visualization.}
    All datasets are encoded by the same unsupervised body-motion encoder and projected into one shared t-SNE space.
    (a) Global projection with balanced sampled points per dataset.
    (b) Support-envelope comparison between \ours and BONES-SEED, the closest large-scale baseline; \ours's support nearly encloses BONES-SEED.
    (c) Grid-based local-coverage statistics on the same embedding using a $55{\times}55$ grid.}
    \label{fig:motion_space_tsne_full}
\end{figure*}

\subsection{Motion-Space Diversity}
\label{sec:motion_space_diversity}

We use a dataset-balanced protocol to compare kinematic motion-space coverage. All datasets are first mapped to a unified 23-keypoint, 30 FPS, root-aligned, and body-scale-normalized representation, and each sequence is divided into one-second windows. For shared-encoder training, we randomly sample 5,000 clips from each dataset and draw one window from each sampled clip. One exception is LAFAN1 that its full dataset contains fewer clips than 5,000, and thus we use its all available clips for comparison. 
We then train a shared temporal-convolutional autoencoder using only a reconstruction objective and project its 16-D latent codes into a common t-SNE space. The visualization and occupancy analysis likewise use at most 5,000 windows per dataset (see Appendix~\ref{app:motion_space_diversity}). Figure~\ref{fig:motion_space_tsne_full} shows that \ours spans the broadest kinematic region among the compared datasets, covering most regions occupied by the other datasets while extending into additional parts of the motion space. We further discretize the embedding and report three statistics (defined in Appendix~\ref{app:motion_space_diversity}): \emph{occupied cells} (how wide the coverage is), \emph{effective occupancy} (how uniform the coverage is across occupied cells), and \emph{long-tail share} (the fraction of samples in globally rare cells). \ours covers more grid cells than the closest baseline BONES-SEED (1620 vs.~1438), with larger effective occupancy (1443 vs.~1114) and higher long-tail share (14.1\% vs.~10.7\%), indicating broader and more uniform coverage rather than denser sampling of common motions. In other words, \ours not only reaches more regions of the kinematic space, but also distributes its motions more broadly across those regions, under the same randomly sampled size. We further repeat the coverage analysis across multiple random seeds, t-SNE perplexities, and grid resolutions, and refer the detailed results to Appendix~\ref{app:motion_space_diversity}. \ours retains a positive occupied-cell margin over the strongest baseline in every tested setting, confirming that its coverage advantage is consistent across projection and discretization choices.

\subsection{Data Quality}
\label{sec:data_quality_precision}

\paragraph{Body-motion precision.}
We measure five quantities (\emph{lower is better}; aggregation defined in Appendix~\ref{app:data_quality_precision}) that capture common failure modes in humanoid imitation: jerk $\mathcal{J}_{\tau}$ (m/s$^3$), acceleration $\mathcal{A}$ (m/s$^2$), ground penetration $\delta_{\text{ground}}$ (mm), unsupported-floating share $\phi_{\text{float}}$ (\%), and support-point drift $\nu_{\text{foot}}$ (mm/s).
$\mathbb{Q}_{\tau}$ and $\mathbb{E}$ denote the upper-tail quantile and mean ($\tau{=}0.95$); $\nabla_t$ is a finite-difference operator divided by the source-specific frame interval $\Delta t$, so that $\nabla_t^k x$ has units of m/s$^k$; $\bar{x}_j$ and $x_j$ are smoothed and raw joint positions; $\mathcal{C}$ is the core-body joint set; $\mathcal{F}$ is the support-point set; $\mathcal{K}=\{(t,f):h_f(t)-g<\epsilon\}$ is the contact set of (frame, support-point) pairs within $\epsilon=30$~mm of the ground at height $g$; $h_f$ and $p_f$ are the height and position of support point $f$; $u(t)$ is the unsupported-floating indicator; and $\Pi_{xy}(\cdot)$ is an operator that projects the input onto the ground plane. Formally, we have
{\small
\begin{equation}
\begin{aligned}
&\mathcal{J}_{\tau}
=\mathbb{Q}_{\tau}\!\left\{\left\|\nabla_t^3\bar{x}_j(t)\right\|_2:j\in\mathcal{C}\right\},
\,\,
\mathcal{A}
=\mathbb{Q}_{1/2}\!\left\{\left\|\nabla_t^2 x_j(t)\right\|_2:j\in\mathcal{C}\right\},
\,\,
\phi_{\text{float}}
=\mathbb{E}[u(t)],\\[-0.5mm]
&\delta_{\text{ground}}
=\mathbb{Q}_{\tau}\!\left\{\max\!\bigl(g-h_f(t),0\bigr):f\in\mathcal{F}\right\},
\quad
\nu_{\text{foot}}
=\mathbb{E}\!\left[\left\|\Pi_{xy}\left(\nabla_t p_f(t)\right)\right\|_2:(t,f)\in\mathcal{K}\right].
\end{aligned}
\label{eq:body_quality}
\end{equation}
}

Floor-related metrics ($\delta_{\text{ground}}$, $\phi_{\text{float}}$, $\nu_{\text{foot}}$) require a comparable absolute ground convention. Motion-X++ does not provide one and is therefore excluded from these three metrics in Table~\ref{tab:sec52_quality_compact}(a).

\begin{table*}[t]
\centering
\caption{\textbf{Data quality comparison.}
(a) Body-motion smoothness and ground-contact metrics, lower is better: jerk $\mathcal{J}_{\tau}$, acceleration $\mathcal{A}$, below-ground depth $\delta_{\mathrm{ground}}$, unsupported-floating share $\phi_{\mathrm{float}}$, and support-point drift $\nu_{\mathrm{foot}}$.
(b) Human-object geometric consistency, higher is better: non-conflict fraction $\eta_{\mathrm{nc}}$ and near-surface grounding within 20\,cm $\rho_{\mathrm{near}}$.}
\vspace{0.1in}
\label{tab:sec52_quality_compact}
\scriptsize
\setlength{\tabcolsep}{1.55pt}
\renewcommand{\arraystretch}{1.05}

\begin{minipage}[t]{0.60\textwidth}
\centering
\textbf{(a) Body motion precision}\vspace{0.5mm}

\begin{tabular*}{\linewidth}{@{\extracolsep{\fill}}lrrrrrr@{}}
\toprule
\textbf{Dataset}
& \shortstack{\textbf{Duration}\\\textbf{(h)}}
& \shortstack{$\bm{\mathcal{J}}_{\tau}$\\(m/s$^3$)}
& \shortstack{$\bm{\mathcal{A}}$\\(m/s$^2$)}
& \shortstack{$\bm{\delta}_{\mathrm{ground}}$\\(mm)}
& \shortstack{$\bm{\phi}_{\mathrm{float}}$\\(\%)}
& \shortstack{$\bm{\nu}_{\mathrm{foot}}$\\(mm/s)} \\
\midrule
\textbf{\ours} & \textbf{617.5} & \textbf{173.9} & \textbf{10.7} & \textbf{8} & \textbf{0.015} & \textbf{64} \\
AMASS & 62.9 & 529.0 & 14.9 & 111 & 0.76 & 87 \\
BONES-SEED & 288.3 & 294.2 & 11.3 & 18 & 1.02 & 86 \\
Motion-X++$^{\dagger}$ & 26.0 & 709.0 & 23.8 & -- & -- & -- \\
LAFAN1 & 4.6 & 383.1 & 23.4 & 29 & 7.22 & 157 \\
\bottomrule
\end{tabular*}
\end{minipage}
\hfill
\begin{minipage}[t]{0.37\textwidth}
\centering
\textbf{(b) Object interaction geometry}\vspace{0.5mm}

\begin{tabular*}{\linewidth}{@{\extracolsep{\fill}}lrrr@{}}
\toprule
\textbf{Dataset}
& \shortstack{\textbf{Duration}\\\textbf{(h)}}
& \shortstack{$\bm{\eta}_{\mathrm{nc}}$\\(\%)}
& \shortstack{$\bm{\rho}_{\mathrm{near}}$\\(\%)} \\
\midrule
\textbf{\ours} & \textbf{245.7} & 98.1 & \textbf{95.7} \\
HIMO$^{\ast}$ & 21.6 & 97.6 & 79.0 \\
OMOMO & 9.8 & 90.6 & 50.4 \\
HUMOTO$^{\ddagger}$ & 0.81 & \textbf{99.9} & 82.9 \\
\bottomrule
\end{tabular*}
\end{minipage}

\vspace{-2mm}
\begin{flushleft}
\footnotesize
\vspace{0.1in}
All metrics are aggregated over the full evaluated duration.
$^{\ast}$HIMO contains 9.4~h of unique multi-object sequences. 
Since each object in a multi-object sequence is evaluated separately,
the reported duration becomes 21.6 object-track hours.
$^{\dagger}$Motion-X++ floor metrics require a comparable absolute floor convention and are therefore omitted.
$^{\ddagger}$HUMOTO is evaluated on its public subset; the full licensed dataset is not directly downloadable as public data.
\end{flushleft}
\vspace{-1mm}
\end{table*}

Among datasets sharing the same floor-plane convention, \ours achieves the best value on every reported body-motion metric in Table~\ref{tab:sec52_quality_compact}(a). These metrics correspond to important physical failure modes in humanoid imitation: $\mathcal{J}_{\tau}$ measures abrupt motion changes, $\mathcal{A}$ captures excessive acceleration, $\delta_{\text{ground}}$ indicates ground penetration, $\phi_{\text{float}}$ reflects unsupported floating, and $\nu_{\text{foot}}$ measures foot sliding during contact.

\paragraph{Object-interaction consistency.}
Two quantities (\emph{higher is better}) measure whether human motion and object geometry remain physically consistent in a shared frame: non-conflict fraction $\eta_{\text{nc}}$ (\%) and near-surface grounding $\rho_{\text{near}}$ (\%). Let $\mathcal{S}(t)$ be a set of points sampled along the human-skeleton segment centerlines (uniformly along each segment), $\Omega(t)$ the posed object as a closed 3D region, and $d(\mathcal{S},\partial\Omega)$ the minimum point-to-surface distance from $\mathcal{S}$ to the object boundary $\partial\Omega$. Below, $\mathbf{1}[\cdot]$ is the indicator function (1 if the condition holds, 0 otherwise):
{\small
\begin{equation}
\eta_{\text{nc}}=\mathbb{E}\!\bigl[\mathbf{1}[\,\mathcal{S}(t)\cap\Omega(t)=\emptyset\,]\bigr],
\qquad
\rho_{\text{near}}=\mathbb{E}\!\bigl[\mathbf{1}[\,d(\mathcal{S}(t),\partial\Omega(t))<0.20\,\text{m}\,]\bigr].
\label{eq:object_ratios}
\end{equation}
}

Table~\ref{tab:sec52_quality_compact}(b) reports $\eta_{\text{nc}}$ and $\rho_{\text{near}}$. \ours provides substantially longer object-interaction duration than existing full-body HOI datasets (245.7~h vs.\ 21.6~h for HIMO and 9.8~h for OMOMO) while maintaining strong geometric consistency. In particular, \ours achieves 98.1\% non-conflict rate and 95.7\% near-surface grounding, substantially improving interaction grounding over HIMO (79.0\%) and OMOMO (50.4\%).

\subsection{Humanoid Learning}
\label{sec:humanoid tracking}

We evaluate the physical executability of \ours through physics-based humanoid tracking. For whole-body motion, we compare \ours with AMASS~\citep{mahmood2019amass}, LAFAN1~\citep{harvey2020robust}, Motion-X++~\citep{zhang2025motion}, and BONES-SEED~\citep{bonesseed2026}. All motions are retargeted to the Unitree G1 robot and trained with the same DeepMimic imitation learning pipeline~\citep{peng2018deepmimic}. For motion-with-object tracking, we compare with HUMOTO~\citep{lu2025humoto} and OMOMO~\citep{li2023omomo}, using Omni-Retarget~\citep{yang2025omniretarget} and BeyondMimic-style~\citep{liao2025beyondmimic} to handle synchronized human-object demonstrations. This unified setup allows performance differences to mainly reflect the usability of the motion data. All experiments are conducted from unmirrored original motions.

\subsubsection{Whole-Body Motion Tracking}
\label{sec:whole body motion}
Whole-body tracking tests how well each motion source can be physically imitated without the additional complexity of object interaction. Due to the limited usable duration of LAFAN1, we construct a balanced 3-hour subset for each data source and additionally evaluate a 20-hour setting for larger datasets, to ensure a fair comparison independent of HiPHI's data-scale advantage. For each setting, subsets are sampled using the same random sampling procedure, and all policies are trained under the same optimization budget, with additional details provided in Appendix~\ref{app:whole_body_training_details}. A rollout is considered successful only if its mean full-body position error remains below 0.5~m at every simulation step; otherwise, it is marked as failed.

\begin{figure}[t]
\centering
\includegraphics[width=0.95\linewidth]{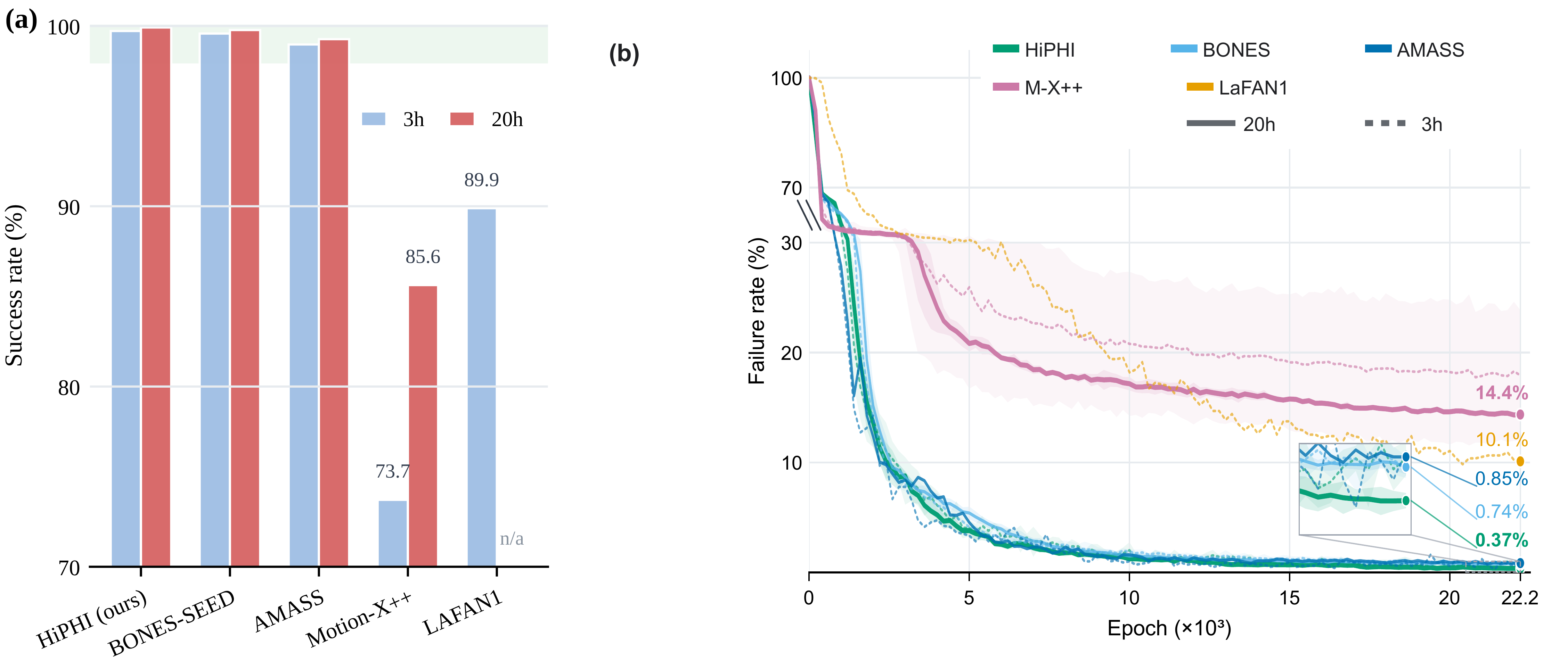}
\caption{
\textbf{Whole-body humanoid tracking performance across different motion data sources.}
(a) Success rates under matched 3-hour and 20-hour data budgets.
(b) Training failure-rate curves averaged over five independent runs, with shaded regions indicating variance.
\ours achieves faster convergence and lower final failure rates across different data scales.
}
\label{fig:success_rate}
\end{figure}

As shown in Figure~\ref{fig:success_rate}, \ours achieves the highest success rates and fastest convergence in both the 3-hour and 20-hour settings.
The advantage remains consistent across five independent training runs, with \ours maintaining lower failure rates and smaller variance than the compared datasets.
These results show that \ours provides motion references that are not only larger in scale, but also more directly executable for humanoid policy learning.

\subsubsection{Data Scaling}

\begin{figure}[t]
    \centering
    \includegraphics[width=0.95\linewidth]{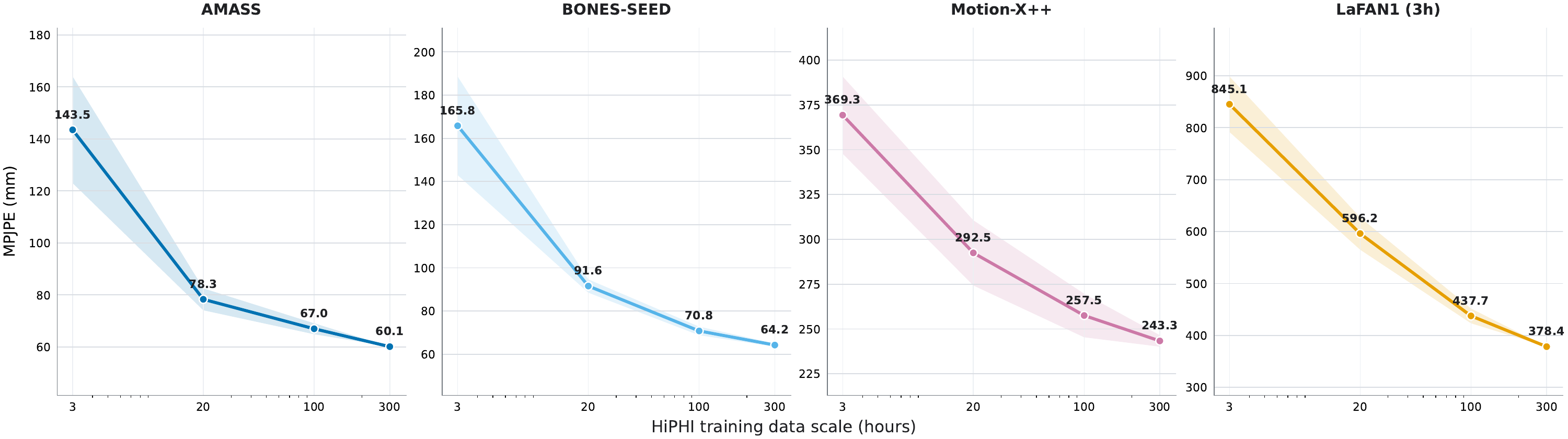}
    \caption{
    \textbf{Scaling behavior of \ours for humanoid tracking.}
    Increasing the amount of unmirrored \ours training data from 3 to 300 hours consistently reduces cross-dataset MPJPE on AMASS, BONES-SEED, Motion-X++, and LAFAN1, demonstrating continuous performance gains with increasing data scale.
    }
    \label{fig:hiphi_scaling}
\end{figure}

Beyond the matched-budget comparison, we further investigate how the scale of \ours affects downstream humanoid tracking.
We progressively increase the amount of unmirrored \ours training data from 3 to 300 hours and evaluate the resulting policies on AMASS, BONES-SEED, Motion-X++, and LAFAN1. Consistent with the experiments above, we randomly sample 20 hours from each dataset as the evaluation set, except for LAFAN1, for which 3-hour setting is used. All experiments are repeated 10 times, and we report the mean curves with variance bands.
As shown in Figure~\ref{fig:hiphi_scaling}, increasing the amount of training data consistently reduces cross-dataset MPJPE, demonstrating that the large scale of \ours translates into continuous downstream performance gains.

\subsubsection{Whole-Body Motion Tracking with Object}
\label{sec:whole body track with obj}
We further evaluate human-object interaction tracking on four categories: \textit{kick}, \textit{carry}, \textit{push}, and \textit{lean}. All datasets are processed using the same retargeting and physics-based tracking pipeline, allowing us to evaluate how effectively each dataset supports physically grounded human-object imitation. Some categories are unavailable in existing datasets: HUMOTO does not contain \textit{push} motions, and OMOMO does not provide \textit{lean} motions. These cases are marked as ``N/A'' in Table~\ref{tab:mwo_tracking_results}. 

\begin{table*}[h]
\centering
\scriptsize
\setlength{\tabcolsep}{3.5pt}
\caption{ \textbf{Whole-body motion tracking with object comparison}
(a) Body tracking errors are measured by MPJPE in mm and Vel. in mm/frame.
(b) Object tracking errors are measured by Obj-Pos. in mm and Obj-Ori. in degrees.
}

\label{tab:mwo_tracking_results}

\makebox[\textwidth][c]{
\begin{tabular*}{0.98\textwidth}{@{\extracolsep{\fill}}lcc|cc|cc|cc@{}}
\toprule
\multicolumn{9}{c}{\textbf{(a) Body tracking errors}} \\
\midrule
\multirow{2}{*}{Dataset}
& \multicolumn{2}{c|}{\textbf{Kick}}
& \multicolumn{2}{c|}{\textbf{Carry}}
& \multicolumn{2}{c|}{\textbf{Push}}
& \multicolumn{2}{c}{\textbf{Lean}} \\
\cmidrule(lr){2-3}
\cmidrule(lr){4-5}
\cmidrule(lr){6-7}
\cmidrule(lr){8-9}
& MPJPE $\downarrow$ & Vel. $\downarrow$
& MPJPE $\downarrow$ & Vel. $\downarrow$
& MPJPE $\downarrow$ & Vel. $\downarrow$
& MPJPE $\downarrow$ & Vel. $\downarrow$ \\
\midrule
HUMOTO~\citep{lu2025humoto}
& 47.77 & 4.75
& 55.37 & 11.67
& N/A & N/A
& 38.83 & 4.67 \\

OMOMO~\citep{li2023omomo}
& 85.31 & 7.05
& 75.57 & 6.83
& \textbf{66.09} & 11.28
& N/A & N/A \\

\ours\textbf{(ours)}
& \textbf{26.31} & \textbf{3.68}
& \textbf{50.33} & \textbf{4.14}
& 99.20 & \textbf{6.26}
& \textbf{30.76} & \textbf{2.12} \\
\bottomrule
\end{tabular*}
}

\vspace{4pt}

\makebox[\textwidth][c]{
\begin{tabular*}{0.98\textwidth}{@{\extracolsep{\fill}}lcc|cc|cc|cc@{}}
\toprule
\multicolumn{9}{c}{\textbf{(b) Object tracking errors}} \\
\midrule
\multirow{2}{*}{Dataset}
& \multicolumn{2}{c|}{\textbf{Kick}}
& \multicolumn{2}{c|}{\textbf{Carry}}
& \multicolumn{2}{c|}{\textbf{Push}}
& \multicolumn{2}{c}{\textbf{Lean}} \\
\cmidrule(lr){2-3}
\cmidrule(lr){4-5}
\cmidrule(lr){6-7}
\cmidrule(lr){8-9}
& Obj-Pos. $\downarrow$ & Obj-Ori. $\downarrow$
& Obj-Pos. $\downarrow$ & Obj-Ori. $\downarrow$
& Obj-Pos. $\downarrow$ & Obj-Ori. $\downarrow$
& Obj-Pos. $\downarrow$ & Obj-Ori. $\downarrow$ \\
\midrule
HUMOTO~\citep{lu2025humoto}
& 214.91 & 35.92
& \textbf{59.20} & 5.39
& N/A & N/A
& 136.71 & 4.81 \\

OMOMO~\citep{li2023omomo}
& 128.62 & 104.28
& 62.46 & 4.87
& 257.82 & 46.24
& N/A & N/A \\

\ours\textbf{(ours)}
& \textbf{68.17} & \textbf{3.27}
& 89.13 & \textbf{2.64}
& \textbf{60.51} & \textbf{2.23}
& \textbf{69.16} & \textbf{9.63} \\
\bottomrule
\end{tabular*}
}

\vspace{2pt}
\end{table*}

We report both body-level and object-level tracking errors. Mean per-joint position error~(MPJPE) and velocity error~(Vel.) measure humanoid pose and motion tracking quality, while object position error (Obj-Pos.) and object orientation error~(Obj-Ori.) measure object trajectory and rotation accuracy. As shown in Table~\ref{tab:mwo_tracking_results}, \ours achieves the best overall body tracking performance across most categories. For object tracking, \ours obtains the best position and orientation accuracy on \textit{kick} and \textit{push}, and the best orientation accuracy on \textit{carry}, indicating better interaction consistency between the humanoid and manipulated objects. Additional experimental details are provided in Appendix~\ref{app:mwo_training_details}.

\subsection{Real-World Evaluation of \ours}
\label{sec:real world}

\begin{figure}[h]
    \centering
    \includegraphics[width=0.98\linewidth]{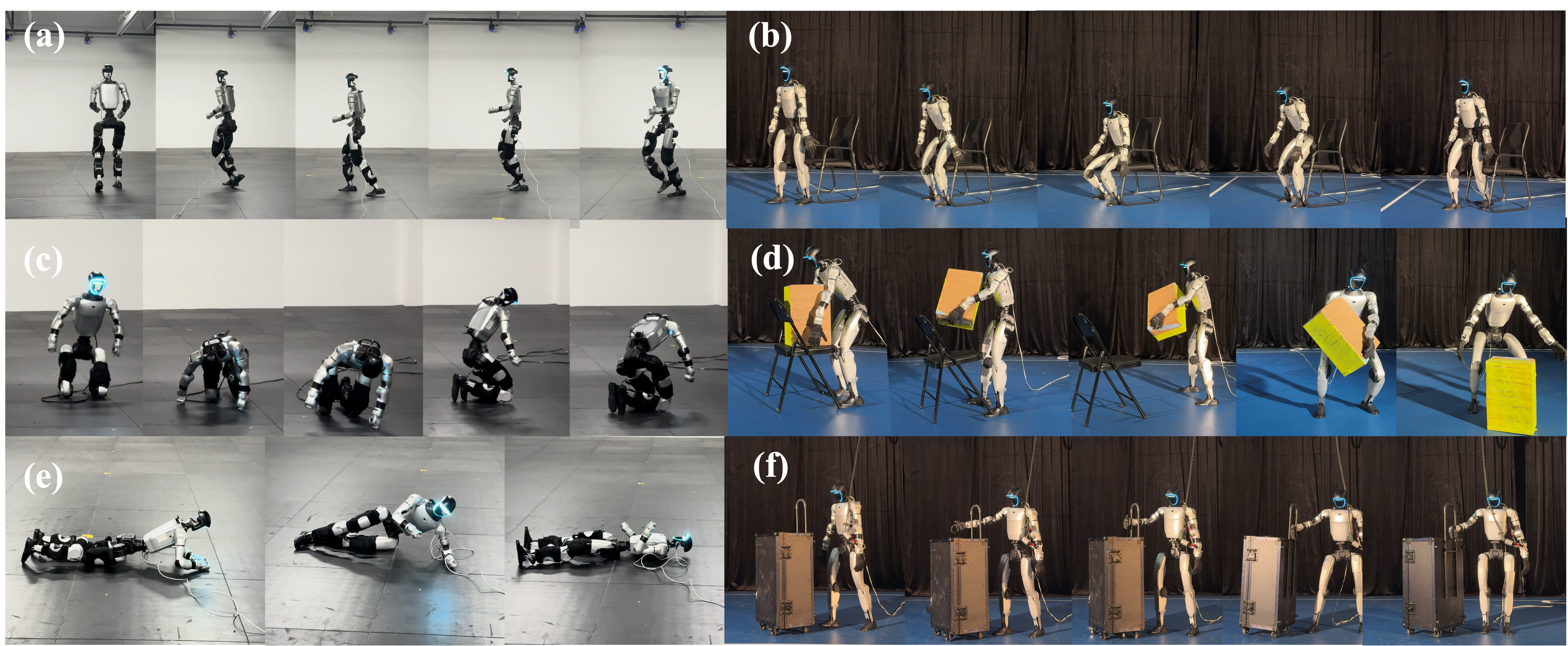}
    \caption{
    \textbf{Real-world deployment of policies trained with \ours.}
    The humanoid successfully performs diverse whole-body and object-interaction behaviors, including (a)~running, (b)~sitting, (c)~crawling, (d)~carrying a box, (e)~flipping, and (f)~pulling a suitcase, showing the physical executability of our motion data.
    }
    \label{fig:real_world_qualitative}
\end{figure}

To evaluate the practical usability of \ours beyond simulation, we deploy policies trained with \ours on Unitree G1. Despite actuation limits, sensing noise, and sim-to-real discrepancies, the robot successfully performs diverse behaviors, including locomotion, sitting, crawling, carrying, flipping, and pulling motions. As shown in Figure~\ref{fig:real_world_qualitative}, these results demonstrate that \ours supports stable and physically plausible whole-body and interaction-rich control on real hardware.

\section{Conclusion}
\label{sec:conclusion}

We introduced \ours, a 617.5-hour high-precision optical MoCap dataset and benchmark for humanoid learning, including 245.7 hours of synchronized human-object interaction with object trajectories and meshes. Built upon a Frame-LU-guided motion-space construction pipeline, \ours provides a systematic and scalable approach for collecting broad whole-body motion and physically grounded interactions.

Through comprehensive evaluations, \ours demonstrates broader motion-space coverage, higher motion quality, stronger human-object geometric consistency, and improved downstream humanoid tracking performance. Policies trained with \ours achieve strong matched-budget tracking performance, continue to improve with increasing data scale, and successfully transfer to real humanoid hardware across diverse whole-body and object-interaction behaviors.

Together, these results establish \ours as a large-scale data foundation for learning physically grounded humanoid skills, bridging high-precision motion capture, object-aware interaction, and scalable policy learning.
\section{Limitations}
\label{sec:limitations}


\ours currently captures only single-person motion. Multi-person interaction and human-human contact remain outside its scope and require complementary efforts.
In addition, although \ours provides high-precision full-body motion and object trajectories, it focuses on kinematics rather than directly measuring contact forces or tactile signals. 
Finally, the dataset is collected in studio, and capturing accurate motion and object interaction with ego-vision in-the-wild could be a promising future data collection direction. 

\acknowledgments{
We sincerely thank Roch Nakajima for generously sharing his extensive experience in motion capture and for his invaluable support in interpreting FrameNet lexical units and assessing their suitability for motion capture. We thank Alex Martinez for developing a practical data visualization tool that greatly facilitated data inspection and quality control. We also thank Baoze Du for his valuable advice on standardizing the BVH skeleton specification. Finally, we thank the many other staff members at Noitom Robotics who operated and maintained the motion-capture facilities and equipment, coordinated data-collection logistics and infrastructure, and supported the performers throughout the capture process. We are also deeply grateful to all performers who contributed their time and effort to the creation of HiPHI.}

\bibliography{example}

\clearpage
\clearpage
\appendix
\section*{Appendix}

The appendix provides additional details that complement the main text. Appendix~\ref{app:data_construction}-\ref{app:dataset_composition_metadata} describe FrameNet-guided motion-unit construction and the release composition, metadata, and demographics. Appendix~\ref{app:release_ethics} states the release license and ethics. Appendix~\ref{app:samples} shows representative sample sequences. Appendix~\ref{app:motion_space_diversity}-\ref{app:data_quality_precision} provide the unsupervised motion-space embedding setup and the full-run quality metrics. Appendix~\ref{app:whole_body_training_details}-\ref{app:mwo_training_details} document training details for whole-body and motion-with-object tracking.

\section{Additional Details for Dataset Construction}
\label{app:data_construction}

\paragraph{FrameNet-guided motion units.}
\ours uses FrameNet as a semantic scaffold rather than a closed action taxonomy. We identify motion-relevant frames and lexical units (LUs), and use each Frame-LU pair as a seed for a motion unit that can be translated into performer-facing capture instructions. This keeps the public index compact while allowing each semantic seed to expand into multiple concrete realizations.

\begin{table}[h]
\centering
\caption{Examples of FrameNet-guided motion-unit construction. Frame-LU seeds are expanded by controlled realization factors rather than treated as single fixed actions.}
\label{tab:framenet_realization_examples}
\small
\setlength{\tabcolsep}{4pt}
\renewcommand{\arraystretch}{1.08}
\begin{tabular}{p{0.22\linewidth}p{0.27\linewidth}p{0.42\linewidth}}
\toprule
\textbf{Motion family} & \textbf{Frame-LU examples} & \textbf{Realization factors} \\
\midrule
Locomotion and direction & \emph{Self\_motion}: walk, run, jog, stride & path, direction, speed, turning radius, step rhythm \\
Posture and support & \emph{Change\_posture}, \emph{Posture}: crawl, kneel, lean & body height, support relation, amplitude, transition speed \\
Body-part motion & \emph{Body\_movement}: bend, toss, shake, clap & body part, range of motion, rhythm, unilateral/bilateral pattern \\
Object actuation and transfer & \emph{Cause\_motion}, \emph{Bringing}: push, pull, lift, carry & object category, contact mode, load condition, object trajectory \\
Object-local dynamics & \emph{Cause\_to\_move\_in\_place}: rotate, shake, swing & axis of motion, contact point, amplitude, repetition pattern \\
\bottomrule
\end{tabular}
\end{table}

\paragraph{Performer-facing realization protocol.}
Each selected Frame-LU seed is converted into short, repeatable capture instructions by varying factors that are directly observable during MoCap: direction, path, speed, intensity, amplitude, body height, body-part involvement, object category, contact mode, and load condition. For instance, a \emph{push} seed is realized with different object categories, loads, and directions; a \emph{lean} seed varies support surface and body height; and a \emph{crawl} seed varies path shape, speed, and limb coordination. This protocol turns a compact semantic index into a broader set of controllable whole-body motion realizations.

\section{Additional Details for Dataset Composition and Metadata}
\label{app:dataset_composition_metadata}

\paragraph{Release duration convention.}
All release-scale durations in the paper follow the public mirrored-release convention, where left-right mirrored sequences are included as normal released data. Under this convention, \ours contains 617.5 hours of motion, corresponding to approximately 200.1M frames at 90 Hz. The release contains 371.8 hours of body-only motion and 245.7 hours of strict object-interaction motion, where strict object-interaction denotes explicit physical interaction with a tracked object.

\begin{table}[t]
\centering
\caption{Release-level composition used throughout the paper. Durations follow the mirrored-release convention.}
\label{tab:release_composition_appendix}
\small
\setlength{\tabcolsep}{5pt}
\renewcommand{\arraystretch}{1.08}
\begin{tabular}{l r}
\toprule
\textbf{Statistic} & \textbf{Value} \\
\midrule
Total release duration & 617.5 h \\
Total frames & 200.1M \\
Capture frame rate & 90 Hz \\
Total performers & 132 \\
Body-only duration & 371.8 h \\
Human object-interaction duration & 245.7 h \\
Human object-interaction share & 39.8\% \\
FrameNet frames & 22 \\
Frame-LU labels & 214 \\
Median actors per Frame-LU & 24 \\
Frame-LUs with at least 10 actors & 154 \\
Object categories in strict object-interaction subset & 12 \\
\bottomrule
\end{tabular}
\end{table}

\paragraph{Frame-LU composition.}
The released semantic layer contains 22 FrameNet frames and 214 Frame-LU motion-unit labels. Its duration distribution follows a common-to-long-tail structure: the top 10 Frame-LUs account for 20.4\% of duration, the top 20 for 31.8\%, and the top 50 for 53.7\%. Thus, nearly half of the release duration remains outside the top 50 Frame-LUs, supporting retrieval and analysis beyond a small set of frequent actions.

\begin{figure}[t]
    \centering
    \includegraphics[width=\linewidth]{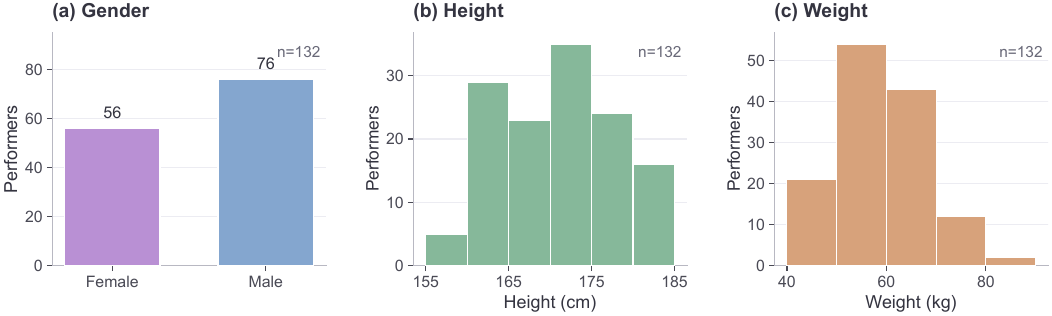}
    \caption{\textbf{Performer demographics of \ours.}
    (a) Gender split among 132 performers.
    (b) Height distribution.
    (c) Weight distribution.}
    \label{fig:app_actor_demographics}
\end{figure}

\paragraph{Performer demographics.}
\ours contains 132 anonymized performer profiles,
including 76 male and 56 female profiles, with heights from 155 to 185~cm and weights from 40 to 82~kg.Figure~\ref{fig:app_actor_demographics} summarizes the distribution.

\begin{table}[t]
\centering
\caption{
\textbf{Metadata fields exposed by the HiPHI release.}
Fields are provided through the global motion manifest, performer metadata,
and per-package \texttt{metadata.json} files.
}
\label{tab:release_metadata_fields}
\scriptsize
\setlength{\tabcolsep}{3.2pt}
\renewcommand{\arraystretch}{1.02}

\begin{tabularx}{\linewidth}{
@{}
>{\raggedright\arraybackslash}p{0.30\linewidth}
>{\raggedright\arraybackslash}p{0.18\linewidth}
>{\raggedright\arraybackslash}X
@{}}
\toprule
\textbf{Field} & \textbf{Scope} & \textbf{Meaning} \\
\midrule

\multicolumn{3}{@{}l}{\textbf{Global motion metadata}} \\[0.5mm]

\texttt{motion\_id}
& Global / package
& Unique motion identifier. Mirrored motions use the corresponding ID with a \texttt{\_\_mirror} suffix. \\

\texttt{frame}
& Global / package
& FrameNet frame label. \\

\texttt{lu}
& Global / package
& FrameNet lexical-unit label. \\

\texttt{frame\_lu}
& Global / package
& Frame-LU motion-unit label. \\

\texttt{duration\_sec}
& Global / package
& Temporal duration of the motion, in seconds. \\

\texttt{frame\_count}
& Global / package
& Number of motion frames in the sequence. \\

\texttt{actor\_id}
& Global / package
& Anonymized performer identifier in the release. \\

\texttt{text\_annotation}
& Global / package
& Natural-language annotation aligned with the motion sequence. \\

\texttt{is\_hoi}
& Global / package
& Whether the sequence contains tracked human-object interaction. \\

\texttt{object\_categories}
& Global
& Object-category information associated with the sequence, when applicable. \\

\texttt{mirrored}
& Global / package
& Boolean indicating whether the motion is the left-right mirrored counterpart of an original sequence. \\

\midrule
\multicolumn{3}{@{}l}{\textbf{Performer metadata}} \\[0.5mm]

\texttt{actor\_id}
& Actor metadata
& Anonymized performer identifier used to link performer metadata to motion sequences. \\

\texttt{height\_cm}
& Actor metadata
& Performer height in centimeters. \\

\texttt{weight\_kg}
& Actor metadata
& Performer weight in kilograms. \\

\texttt{gender}
& Actor metadata
& Performer gender metadata. \\

\midrule
\multicolumn{3}{@{}l}{\textbf{Additional per-package metadata}} \\[0.5mm]

\texttt{dataset}
& Package only
& Dataset identifier stored in the package metadata. \\

\texttt{fps}
& Package only
& Frame rate associated with the released motion package. \\

\texttt{actor\_metadata}
& Package only
& Nested performer metadata associated with the sequence. \\

\texttt{objects}
& Package only
& List of tracked objects associated with the sequence; empty for body-only motion. \\

\texttt{objects[].object\_id}
& Package only
& Identifier of a tracked object instance. \\

\texttt{objects[].object\_category}
& Package only
& Semantic category of the tracked object. \\

\texttt{objects[].trajectory\_path}
& Package only
& Package-relative path to the synchronized object trajectory. \\

\texttt{objects[].mesh\_id}
& Package only
& Identifier of the corresponding object mesh. \\

\texttt{objects[].mesh\_path}
& Package only
& Package-relative path to the corresponding object mesh. \\

\bottomrule
\end{tabularx}

\vspace{-1mm}
\end{table}

\paragraph{Object-interaction metadata.} 
The \ours object-interaction subset contains 40 real-world objects from 12 categories, covering a mass range of 0.45-6.25 kg and spanning 15 FrameNet frames and 90 Frame-LU labels (Table~\ref{tab:object_metadata}).
For strict object-interaction sequences, the release provides synchronized object trajectories and object meshes together with the human BVH motion.

\begin{table}[t]
\centering
\caption{
Object statistics of the \ours human-object interaction subset.
The subset contains diverse real-world objects with synchronized human motion, object trajectories, and object meshes.
}
\label{tab:object_metadata}
\small
\begin{tabular}{lc}
\toprule
\textbf{Statistic} & \textbf{Value} \\
\midrule
Object instances & 40 \\
Object categories & 12 \\
Mass range & 0.45-6.25 kg \\
Covered FrameNet frames & 15 \\
Covered Frame-LU labels & 90 \\
Interaction duration & 245.7 h \\
Object state modalities & Trajectory + Mesh \\
\bottomrule
\end{tabular}
\end{table}

\section{Release Terms, License, and Ethics}
\label{app:release_ethics}

\paragraph{License.}
\ours will be released under the
\textbf{ModalityNet Open Research License v1.0},
a custom non-commercial license for scientific research, education, and evaluation.
The complete license terms will accompany the public release of \ours.

\paragraph{Release schedule.}
The full dataset, including BVH motions, object trajectories, object meshes, and the Frame-LU index, will be released publicly soon. The supplementary material contains four representative sample sequences (three body-only and three object-interaction) that illustrate the release format and quality.

\paragraph{Ethics.}
Data collection was reviewed and approved by an institutional ethics review board before capture. All performers were adults who provided written informed consent describing the capture protocol, the intended research use, and the planned public release of the motion data and anonymized demographic attributes. Performers were compensated for each capture session and may request withdrawal of their data at any time. During capture, RGB video and audio were recorded as references for production review; \textbf{the public release of \ours contains only BVH motion, object trajectories, and object meshes, and does not include any RGB video, audio, facial imagery, or voice recordings}. Performer identities are released only as numeric actor identifiers, with anonymized height, weight, and gender attributes. Specific institutional and protocol identifiers are withheld for double-blind review and will be provided in the camera-ready version.

\section{Representative Sample Sequences}
\label{app:samples}

We show six representative sequences from \ours. Per-sequence renderings are shown in Figure~\ref{fig:samples}, and full motion playback is provided in the supplementary video.

\begin{figure*}[h]
    \centering
    \includegraphics[
        width=\textwidth,
        trim=100pt 100pt 0pt 80pt,
        clip
    ]{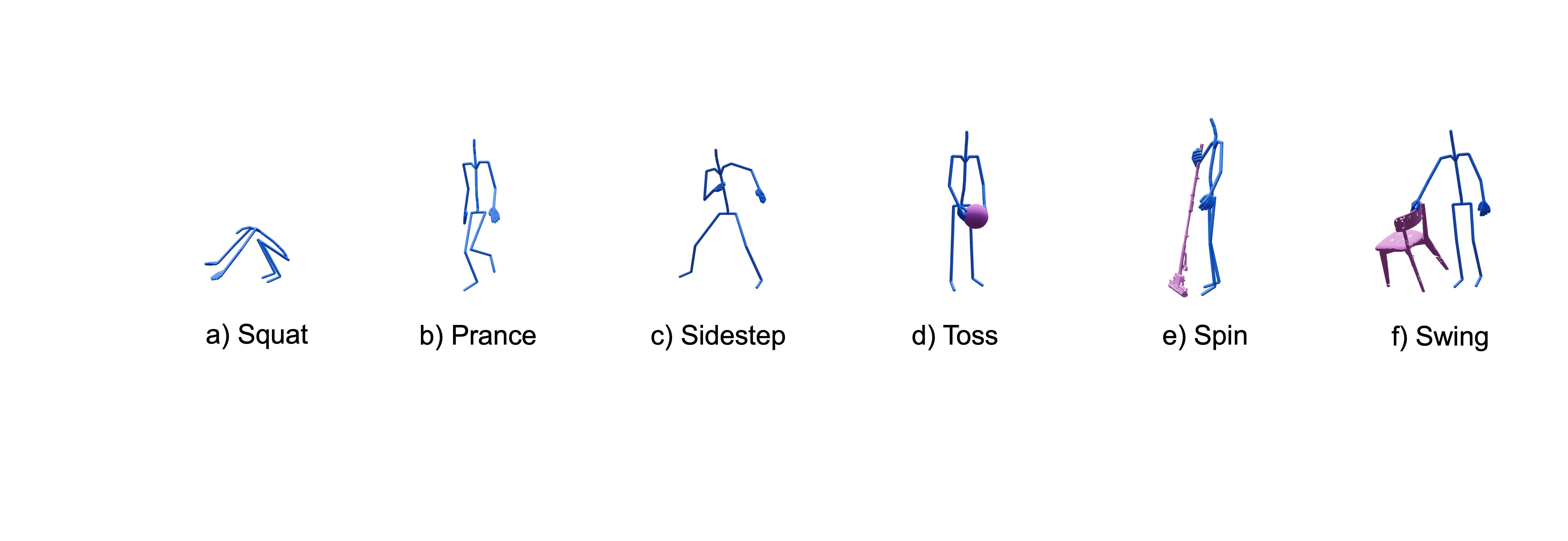}
    \caption{\textbf{Representative sample sequences from \ours.}
    We show six representative sequences from \ours:
    (a)-(c) body-only motion sequences and
    (d)-(f) object-interaction motion sequences.}
    \label{fig:samples}
\end{figure*}

\section{Additional Details for Motion-Space Diversity}
\label{app:motion_space_diversity}

\paragraph{Unified motion representation.}
For the motion-space analysis in Sec.~\ref{sec:motion_space_diversity}, all datasets are first mapped to a common body representation with 23 keypoints in meters, resampled to 30~FPS in a $z$-up coordinate frame.
The keypoints are pelvis, spine, chest, neck, head, left/right hip, knee, ankle, foot, toe, shoulder, elbow, wrist, and hand.
Each sequence is divided into 30-frame windows with a stride of 25 frames.
The per-frame input contains heading-local, root-relative joint positions and root-relative joint velocities.
We further apply body-scale normalization and subtract the per-window median pose, which reduces scale and skeleton-layout differences across datasets.
The resulting feature dimension is $23\times3\times2=138$ per frame.

\paragraph{Dataset-balanced sampling.}
To prevent dataset size from dominating either representation learning or coverage evaluation, we construct a balanced sample for each dataset.
We randomly sample 5{,}000 motion clips from each dataset when available and draw one valid 30-frame window from each selected clip, yielding 5{,}000 training windows per dataset.
LAFAN1 uses all available clips because it contains fewer than 5{,}000 samples.
The t-SNE projection and grid-based occupancy analysis follow the same cap of at most 5{,}000 windows per dataset.
Therefore, both the shared representation and the reported coverage statistics are dataset-balanced rather than proportional to raw dataset size.

\paragraph{Shared unsupervised encoder.}
We train one temporal convolutional autoencoder jointly on the balanced samples from all datasets and use its encoder as the motion descriptor.
The model takes a tensor of size $30\times138$, uses temporal Conv1D layers with residual dilated blocks, and outputs a 16-D latent code.
It is trained only with reconstruction loss; dataset identity, text labels, Frame-LU labels, and object labels are not used.
The default analysis uses the epoch-20 checkpoint.
Before t-SNE, latent features are standardized and reduced to 5 dimensions with PCA.
We then run t-SNE with Euclidean distance, perplexity 30, PCA initialization, automatic learning rate, 1000 iterations, and random seed 42.

\begin{table}[t]
\centering
\caption{Embedding setup for the motion-space coverage analysis.}
\label{tab:app_embedding_setup}
\small
\setlength{\tabcolsep}{4pt}
\renewcommand{\arraystretch}{1.08}
\begin{tabular}{@{}p{0.30\linewidth}p{0.62\linewidth}@{}}
\toprule
\textbf{Item} & \textbf{Setting} \\
\midrule
Keypoints
& 23 body keypoints, meter scale, $z$-up \\

Frame rate
& 30 FPS \\

Window / stride
& 30 frames / 25 frames \\

Feature type
& Heading-local normalized joint dynamics and relative velocities \\

Feature dimension
& 138 per frame \\

Balanced sampling
& Up to 5{,}000 clips and 5{,}000 windows per dataset \\

Encoder
& Shared unsupervised temporal Conv1D autoencoder \\

Latent dimension
& 16 \\

Objective
& Reconstruction MSE \\

Checkpoint
& Epoch 20 \\

t-SNE
& PCA dim.\ 5, perplexity 30, PCA init., auto learning rate,
1000 iterations, seed 42 \\

Grid statistics
& $55\times55$ grid on the default 2-D embedding \\
\bottomrule
\end{tabular}
\end{table}

\paragraph{Grid-based coverage statistics.}
Let $\mathcal{G}$ be the $55\times55$ grid in the shared t-SNE plane and $n_D(c)$ the number of sampled points from dataset $D$ in cell $c\in\mathcal{G}$.
Let
$\mathcal{G}_D^+ = \{c\in\mathcal{G}:n_D(c)>0\}$
be the cells occupied by $D$.
We report occupied grid cells, entropy-based effective grid cells, and long-tail share:
\begin{equation}
\begin{aligned}
m_D
&=
\lvert\mathcal{G}_D^+\rvert,
\qquad
p_D(c)
=
\frac{n_D(c)}
{\sum_{c'\in\mathcal{G}} n_D(c')},
\\[-0.5mm]
e_D
&=
\exp\!\Bigl(
-\sum_{c\in\mathcal{G}_D^+}
p_D(c)\log p_D(c)
\Bigr),
\qquad
\ell_D
=
\sum_{c\in\mathcal{T}}p_D(c).
\end{aligned}
\label{eq:app_grid_stats}
\end{equation}
Here, $m_D$ is the number of occupied cells, measuring how broadly a dataset spans the shared motion space.
The effective occupancy $e_D$ measures how evenly its samples are distributed across the occupied cells.
The long-tail share $\ell_D$ measures the fraction of samples located in globally rare regions.
We use the convention $0\log0=0$.
The set $\mathcal{T}\subseteq\mathcal{G}$ contains non-empty cells whose global occupancy is at or below the 25th percentile of all non-empty global cells; under the default $55\times55$ setting, this threshold corresponds to 7 points per cell.

\paragraph{Coverage robustness analysis.}
We further test whether the occupied-region advantage depends on a particular random seed, t-SNE perplexity, or grid resolution.
Starting from the default configuration in Table~\ref{tab:app_embedding_setup}, we repeat the analysis using random seeds 55, 73, and 112; perplexities 40, 55, and 65; and grid resolutions of $55\times55$, $88\times88$, and $111\times111$.

For every configuration, we report the occupied-cell margin
\begin{equation}
\Delta \mathrm{Occ}
=
m_{\mathrm{HiPHI}}
-
\max_{D\neq\mathrm{HiPHI}}m_D,
\label{eq:app_coverage_margin}
\end{equation}
where the second term is the occupied-cell count of the strongest baseline under the same configuration.
A positive value means that HiPHI covers more cells than every compared dataset.
Because occupied-cell counts naturally change with grid resolution, the margin is reported separately for each resolution.

\begin{figure*}[t]
    \centering
    \includegraphics[width=0.97\textwidth]{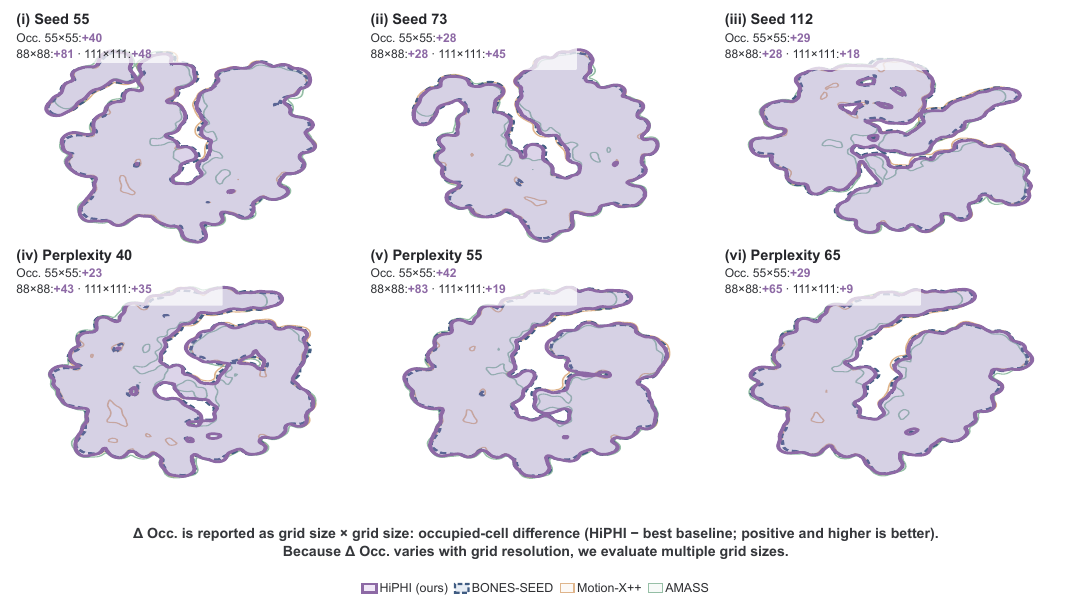}
    \caption{
    \textbf{Robustness of motion-space coverage.}
    We repeat the coverage analysis using different random seeds,
    t-SNE perplexities, and grid resolutions.
    Each entry reports the occupied-cell margin between HiPHI and
    the strongest baseline under the corresponding setting.
    HiPHI maintains a positive margin across all tested configurations.
    }
    \label{fig:tsne_robustness}
\end{figure*}

As shown in Figure~\ref{fig:tsne_robustness}, HiPHI maintains a positive occupied-cell margin in every tested configuration.
The coverage advantage therefore remains consistent across resampling, projection, and discretization choices.

\section{Additional Details for Data Quality and Precision}
\label{app:data_quality_precision}

\paragraph{Aggregation across sequences.}
All quality statistics in Sec.~\ref{sec:data_quality_precision} are aggregated over the full evaluated duration of each dataset, so that every recorded second contributes equally and dataset size does not bias the result. Let $r_k$ index any observation at the relevant level (frame, window, or sequence aggregate) with non-negative duration weight $w_k$ (i.e., $w_k=\Delta t$ of the source for per-frame observations, the window length for per-window observations, and the sequence duration $T_i$ for per-sequence aggregates). We report
\begin{equation}
\mathbb{E}[r]=\frac{\sum_k w_k\, r_k}{\sum_k w_k},
\qquad
\mathbb{Q}_{\tau}(r)=\inf\!\Bigl\{a:\frac{\sum_k w_k\,\mathbf{1}[r_k\le a]}{\sum_k w_k}\ge\tau\Bigr\},
\label{eq:app_duration_aggregation}
\end{equation}
i.e., the time-weighted mean and quantile, with $\tau=0.95$ for upper-tail statistics. For ratio-type metrics ($\eta_{\text{nc}}$, $\rho_{\text{near}}$, $\phi_{\text{float}}$), each $r_i$ is itself a frame-level ratio inside sequence $i$, and weighting by $T_i$ recovers the frame-level aggregation across the dataset. For distribution-type metrics ($\mathcal{J}_{\tau}$, $\mathcal{A}$, $\delta_{\text{ground}}$, $\nu_{\text{foot}}$), $r_k$ is the per-frame (or per-pair) observation entering the corresponding equation in Sec.~\ref{sec:data_quality_precision}.

\paragraph{Body-motion metric sets and smoothing.}
Equation~\ref{eq:body_quality} uses one joint set, one point set, and one pair set. $\mathcal{C}$ is the core-body joint set, excluding fingers and face joints when present. $\mathcal{F}$ is the support-point set: foot, toe, and ankle proxies when available. $\mathcal{K}=\{(t,f):h_f(t)-g<\epsilon\}$ is the contact set of (frame, support-point) pairs within $\epsilon=30$~mm of the ground at height $g$; speed is then used to measure drift over those pairs. The finite-difference operator $\nabla_t$ is divided by the source-specific frame interval $\Delta t$, so $\nabla_t^k x$ has units of m/s$^k$. For jerk, trajectories are smoothed by a fixed physical window of $1/6$~s, implemented as an odd-length moving average with a minimum size of three frames; raw positions are used for acceleration. Floor-related metrics require a comparable absolute ground convention, which is why Motion-X++ is excluded from $\delta_{\text{ground}}$, $\phi_{\text{float}}$, and $\nu_{\text{foot}}$ in Table~\ref{tab:sec52_quality_compact}.

\paragraph{Object-interaction geometry.}
Equation~\ref{eq:object_ratios} uses a sparse skeleton-point proxy. The human body is represented by canonical skeleton segments covering hands, arms, legs, torso, pelvis, and head/neck regions. For each evaluated frame, segment centerlines are uniformly sampled (32 points per segment in the released code) to produce $\mathcal{S}(t)$. The object is represented as a closed 3D region $\Omega(t)$ obtained by posing the released mesh; $\partial\Omega(t)$ is its boundary surface. Distances $d(\mathcal{S},\partial\Omega)$ are computed in the shared coordinate frame. HUMOTO is evaluated on its publicly released GLB subset; the full licensed dataset is not directly downloadable as public data.

\begin{table}[t]
\centering
\caption{Summary of quality metrics used in Sec.~\ref{sec:data_quality_precision}.}
\label{tab:app_quality_metric_summary}
\small
\setlength{\tabcolsep}{3pt}
\renewcommand{\arraystretch}{1.08}
\begin{tabular}{@{}p{0.18\linewidth}p{0.47\linewidth}p{0.16\linewidth}p{0.12\linewidth}@{}}
\toprule
\textbf{Symbol} & \textbf{Meaning} & \textbf{Unit} & \textbf{Better} \\
\midrule
$\mathcal{J}_{\tau}$ & Upper-tail fixed-window jerk over core-body joints & m/s$^3$ & lower \\
$\mathcal{A}$ & Median core-body acceleration & m/s$^2$ & lower \\
$\delta_{\text{ground}}$ & Upper-tail below-ground support-point depth & mm & lower \\
$\phi_{\text{float}}$ & Unsupported-floating duration share & \% & lower \\
$\nu_{\text{foot}}$ & Support-point drift over contact frames & mm/s & lower \\
$\eta_{\text{nc}}$ & Non-conflict fraction under skeleton-point / object-volume proxy & \% & higher \\
$\rho_{\text{near}}$ & Near-surface fraction within 20\,cm of the object surface & \% & higher \\
\bottomrule
\end{tabular}
\end{table}

\paragraph{Applicability.}
The body-motion smoothness metrics require only temporally aligned human joint trajectories and native timing. Ground-contact metrics additionally require a comparable up-axis and absolute ground convention. Object-interaction metrics require synchronized human motion, object pose, and object mesh in a shared coordinate frame. HUMOTO is evaluated on its publicly released GLB subset; the full licensed dataset is not directly downloadable as public data.

\section{Training Details for Whole-Body Motion Tracking}
\label{app:whole_body_training_details}

For whole-body motion tracking, we evaluate all datasets under the same physics-based imitation learning pipeline. All source motions are first retargeted to the Unitree G1 humanoid using PyRoki~\cite{kim2025pyroki}, and the resulting robot motions are then used as reference trajectories for policy learning. We use the same observation design, reward formulation, optimization budget, and evaluation protocol for all data sources so that performance differences mainly reflect the physical usability of the underlying motion data. Each tracking experiment is repeated five times with different random seeds. We report the mean curve and variance across runs. The detailed observation and reward are summarized in Tables~\ref{tab:app_wb_observation} and \ref{tab:app_wb_rewards}.

\begin{figure*}[h]
    \centering
    \includegraphics[width=\textwidth]{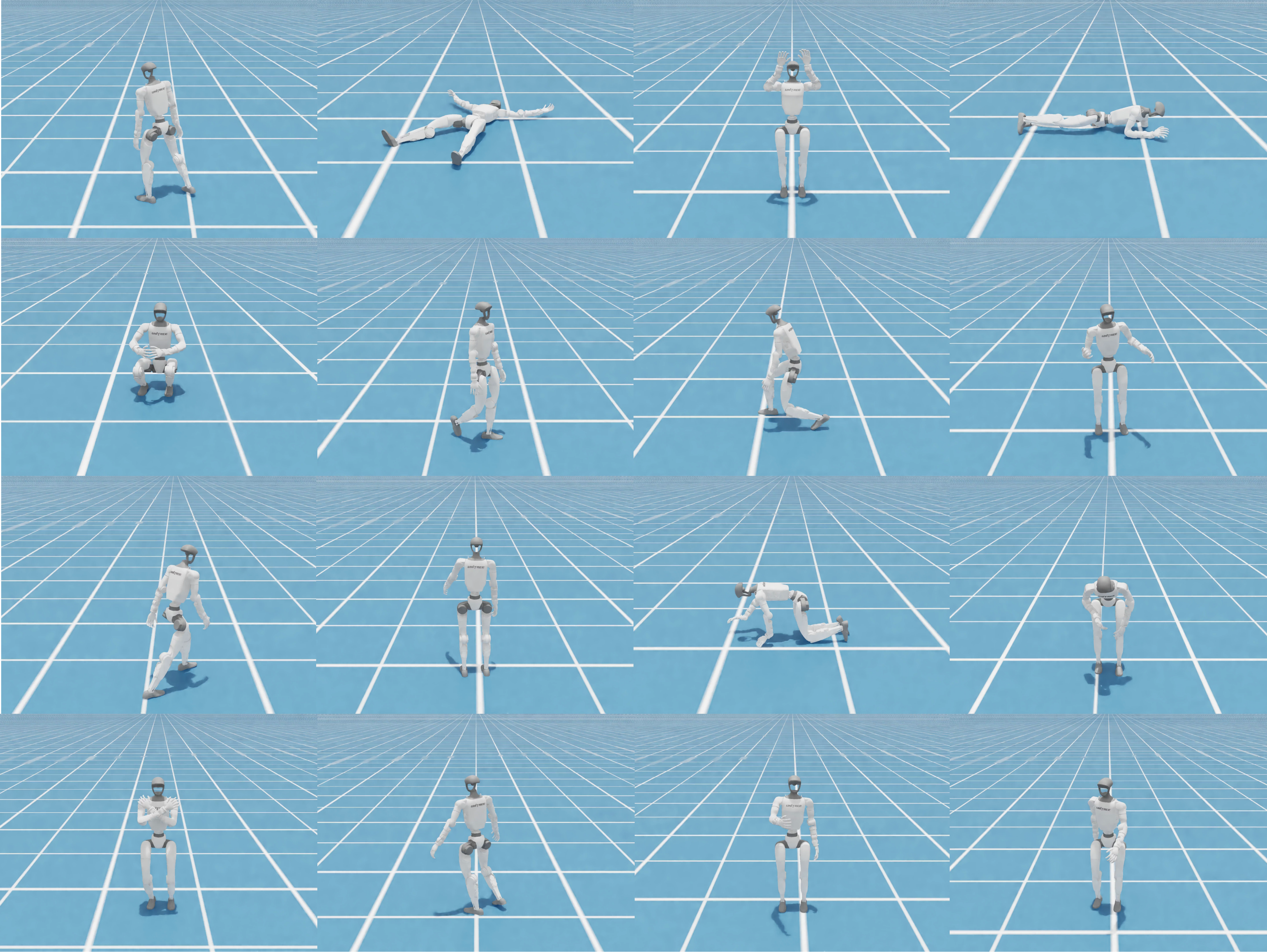}
    \caption{
    \textbf{Qualitative visualization of diverse humanoid motion skills trained with \ours.}
    The motions cover diverse whole-body behaviors, including standing, walking, sitting, lying down, crawling, kneeling, and recovery motions, demonstrating the coverage and diversity of the training data.
}
    \label{fig:wbc_qualitative}
\end{figure*}

Figure~\ref{fig:wbc_qualitative} shows qualitative whole-body tracking results of \ours on the Unitree G1 humanoid. The learned policy produces stable and physically plausible motions across diverse behaviors, demonstrating that the motions in \ours can serve as effective executable references for humanoid control.

\begin{table*}[h]
\centering
\scriptsize
\setlength{\tabcolsep}{4pt}
\renewcommand{\arraystretch}{1.10}
\caption{Observation terms for whole-body motion tracking.}
\label{tab:app_wb_observation}
\begin{tabular*}{0.96\linewidth}{@{\extracolsep{\fill}}lc}
\toprule
\textbf{State} & \textbf{Dim.} \\
\midrule
\rowcolor{gray!15}
\multicolumn{2}{l}{\textbf{(a) Motion Command}} \\
Reference Body Position (heading frame) & $99$ \\
Reference Body Position (relative) & $99$ \\
Reference Body Orientation (heading frame) & $198$ \\
Reference Body Orientation (relative) & $198$ \\
Reference Body Linear Velocity (relative) & $99$ \\
Reference Body Angular Velocity (relative) & $99$ \\
\midrule
\rowcolor{gray!15}
\multicolumn{2}{l}{\textbf{(b) Proprioceptive State}} \\
Root Height (above terrain) & $1$ \\
Body Position (non-root, heading frame) & $96$ \\
Body Orientation (heading frame, tan-norm) & $198$ \\
Body Linear Velocity (heading frame) & $99$ \\
Body Angular Velocity (heading frame) & $99$ \\
Last Action & $29$ \\
\bottomrule
\end{tabular*}
\end{table*}

\begin{table*}[h]
\centering
\scriptsize
\setlength{\tabcolsep}{4pt}
\renewcommand{\arraystretch}{1.12}
\caption{Reward terms used for whole-body motion tracking.}
\label{tab:app_wb_rewards}
\begin{tabular*}{0.98\textwidth}{@{\extracolsep{\fill}}llcl}
\toprule
\textbf{Term} & \textbf{Expression} & \textbf{Weight} & \textbf{Remarks} \\
\midrule
\rowcolor{gray!15}
\multicolumn{4}{c}{\textbf{Tracking Rewards}} \\
Global Translation (gt) &
$\exp\!\left(c_{\mathrm{gt}}\,\frac{1}{n_b}\sum_i \|p_i-\hat{p}_i\|_2^2\right)$
& $0.5$ & $c_{\mathrm{gt}}=-25$ \\
Global Rotation (gr) &
$\exp\!\left(c_{\mathrm{gr}}\,\frac{1}{n_b}\sum_i d_R(q_i,\hat{q}_i)^2\right)$
& $0.3$ & $c_{\mathrm{gr}}=-5$\\
Global Linear Velocity (gv) &
$\exp\!\left(c_{\mathrm{gv}}\,\frac{1}{n_b}\sum_i \|v_i-\hat{v}_i\|_2^2\right)$
& $0.1$ & $c_{\mathrm{gv}}=-0.5$ \\
Global Angular Velocity (gav) &
$\exp\!\left(c_{\mathrm{gav}}\,\frac{1}{n_b}\sum_i \|\omega_i-\hat{\omega}_i\|_2^2\right)$
& $0.2$ & $c_{\mathrm{gav}}=-0.1$ \\
Root Height (rh) &
$\exp\!\left(c_{\mathrm{rh}}\,(z_{\mathrm{root}}-\hat{z}_{\mathrm{root}})^2\right)$
& $0.2$ & $c_{\mathrm{rh}}=-100$ \\
\midrule
\rowcolor{gray!15}
\multicolumn{4}{c}{\textbf{Regularization and Safety Penalties}} \\
Action Smoothness &
$\|\tilde{a}_t-\tilde{a}_{t-1}\|_2$
& $-0.02$ & -- \\
Power Consumption &
$\sum_j |\tau_j\,\dot{q}_j|$
& $-10^{-5}$ & -- \\
\bottomrule
\end{tabular*}
\vspace{1mm}
\begin{minipage}{0.98\textwidth}
\footnotesize
Here $d_R(\cdot,\cdot)$ denotes quaternion orientation error, $n_b$ is the number of tracked body links.
\end{minipage}

\end{table*}



\section{Training Details for Motion-with-Object Tracking}
\label{app:mwo_training_details}

For motion-with-object tracking, we first process synchronized human-object demonstrations using Omni-Retarget~\cite{yang2025omniretarget}, which retargets human motion to the Unitree G1 humanoid while preserving the corresponding object trajectories. Because existing open-source human-object motion datasets usually do not specify object physical parameters, we use a unified object configuration across datasets for fair comparison. Unless otherwise specified, each object is modeled as a free-joint mesh object with a total mass of $2.0$ kg.

We then use a BeyondMimic-style tracking setup~\cite{liao2025beyondmimic}, following a similar observation design and reward formulation while extending the framework to explicitly model object interactions. Since the task requires the humanoid to reproduce both whole-body motion and the associated object trajectory, we additionally include the current object state in the policy observation. We also introduce object-tracking rewards by measuring the position and orientation errors between the simulated object and its reference trajectory. To improve policy robustness, we apply domain randomization to both the humanoid and the object, including object mass, friction, inertial properties, contact parameters, and external perturbations. The detailed observation, reward, and randomization settings are summarized in Tables~\ref{tab:app_mwo_observation}, \ref{tab:app_mwo_rewards}, and \ref{tab:app_mwo_dr}.

\begin{table}[h]
\centering
\scriptsize
\setlength{\tabcolsep}{4pt}
\renewcommand{\arraystretch}{1.10}
\caption{Observation terms for motion-with-object tracking.}
\label{tab:app_mwo_observation}
\begin{tabular*}{0.96\linewidth}{@{\extracolsep{\fill}}lccc}
\toprule
\textbf{State} & \textbf{Dim.} & \textbf{Policy} & \textbf{Critic} \\
\midrule
\rowcolor{gray!15}
\multicolumn{4}{l}{\textbf{(a) Motion Command}} \\
Reference Joint Pos & $29$ & \checkmark & \checkmark \\
Reference Joint Vel & $29$ & \checkmark & \checkmark \\
Reference Anchor Position & $3$ & \checkmark & \checkmark \\
Reference Anchor Orientation & $6$ & \checkmark & \checkmark \\
\midrule
\rowcolor{gray!15}
\multicolumn{4}{l}{\textbf{(b) Proprioceptive State}} \\
Base Linear Velocity & $3$ & \ding{55} & \checkmark \\
Base Angular Velocity & $3$ & \checkmark & \checkmark \\
Joint Position & $29$ & \checkmark & \checkmark \\
Joint Velocity & $29$ & \checkmark & \checkmark \\
Last Action & $29$ & \checkmark & \checkmark \\
\midrule
\rowcolor{gray!15}
\multicolumn{4}{l}{\textbf{(c) Object State}} \\
Object position in robot frame & $3$ & \checkmark & \checkmark \\
Object orientation in robot frame & $6$ & \checkmark & \checkmark \\
\midrule
\rowcolor{gray!15}
\multicolumn{4}{l}{\textbf{(d) Critic-Only Privileged State}} \\
Body position & $14 \times 3$ & \ding{55} & \checkmark \\
Body orientation & $14 \times 6$ & \ding{55} & \checkmark \\
\bottomrule
\end{tabular*}
\end{table}

\begin{table*}[h]
\centering
\scriptsize
\setlength{\tabcolsep}{4pt}
\renewcommand{\arraystretch}{1.12}
\caption{Reward terms used for motion-with-object tracking.}
\label{tab:app_mwo_rewards}
\begin{tabular*}{0.98\textwidth}{@{\extracolsep{\fill}}llcl}
\toprule
\textbf{Term} & \textbf{Expression} & \textbf{Weight} & \textbf{Remarks} \\
\midrule
\rowcolor{gray!15}
\multicolumn{4}{c}{\textbf{Tracking Rewards}} \\
Anchor position &
$\exp\!\left(-\|p_a-\hat{p}_a\|_2^2 / \sigma^2\right)$
& $0.5$ & $\sigma=0.3$ \\
Anchor orientation &
$\exp\!\left(-d_R(q_a,\hat{q}_a)^2 / \sigma^2\right)$
& $0.5$ & $\sigma=0.4$ \\
Body position &
$\exp\!\left(-\frac{1}{n_b}\sum_i \|p_i-\hat{p}_i\|_2^2 / \sigma^2\right)$
& $1.0$ & $\sigma=0.3$ \\
Body orientation &
$\exp\!\left(-\frac{1}{n_b}\sum_i d_R(q_i,\hat{q}_i)^2 / \sigma^2\right)$
& $1.0$ & $\sigma=0.3$ \\
Body linear velocity &
$\exp\!\left(-\frac{1}{n_b}\sum_i \|v_i-\hat{v}_i\|_2^2 / \sigma^2\right)$
& $0.5$ & $\sigma=1.0$ \\
Body angular velocity &
$\exp\!\left(-\frac{1}{n_b}\sum_i \|\omega_i-\hat{\omega}_i\|_2^2 / \sigma^2\right)$
& $0.5$ & $\sigma=3.14$ \\
Object position &
$\exp\!\left(-\|p_o-\hat{p}_o\|_2^2 / \sigma^2\right)$
& $1.0$ & $\sigma=0.3$ \\
Object orientation &
$\exp\!\left(-d_R(q_o,\hat{q}_o)^2 / \sigma^2\right)$
& $1.0$ & $\sigma=0.3$ \\
\midrule
\rowcolor{gray!15}
\multicolumn{4}{c}{\textbf{Regularization and Safety Penalties}} \\
Action rate &
$\|a_t-a_{t-1}\|_2^2$
& $-0.1$ & Smooth action penalty \\
Joint limit &
$\mathbf{1}(q \notin [q_{\min},q_{\max}])$
& $-10.0$ & Joint limit violation \\
\bottomrule
\end{tabular*}

\vspace{1mm}
\begin{minipage}{0.98\textwidth}
\footnotesize
Here $d_R(\cdot,\cdot)$ denotes quaternion orientation error, $n_b$ is the number of tracked body links,
$c$ and $\hat{c}$ denote measured and reference contact states, and hatted variables denote reference motion states.
\end{minipage}
\end{table*}

\begin{table}[h]
\centering
\scriptsize
\setlength{\tabcolsep}{4pt}
\renewcommand{\arraystretch}{1.10}
\caption{Domain randomization and disturbance settings for motion-with-object tracking.}
\label{tab:app_mwo_dr}
\begin{tabular*}{0.96\linewidth}{@{\extracolsep{\fill}}ll}
\toprule
\textbf{Term} & \textbf{Value} \\
\midrule
\rowcolor{gray!15}
\multicolumn{2}{c}{\textbf{External Disturbances}} \\
Push robot &
interval $=1$--$3$s,\quad
$v_x,v_y \sim U[-0.5,0.5]$ m/s,\quad
$v_z \sim U[-0.2,0.2]$ m/s \\
\midrule
\rowcolor{gray!15}
\multicolumn{2}{c}{\textbf{Robot Dynamics Randomization}} \\
Torso COM offset &
$x\sim U[-0.025,0.025]$ m,\quad
$y,z\sim U[-0.05,0.05]$ m \\
Encoder bias &
$U[-0.01,0.01]$ \\
Default joint position offset &
$U[-0.01,0.01]$ rad \\
Rigid-body material &
static friction $U[0.3,1.6]$, dynamic friction $U[0.3,1.2]$, restitution $U[0.0,0.5]$ \\
\midrule
\rowcolor{gray!15}
\multicolumn{2}{c}{\textbf{Object Dynamics Randomization}} \\
Object mass &
$U[0.3,2.0]\times$ default mass \\
Object COM offset &
$U[-0.02,0.02]$ m \\
\bottomrule
\end{tabular*}
\end{table}

\begin{figure*}[h]
    \centering
    \includegraphics[width=\textwidth]{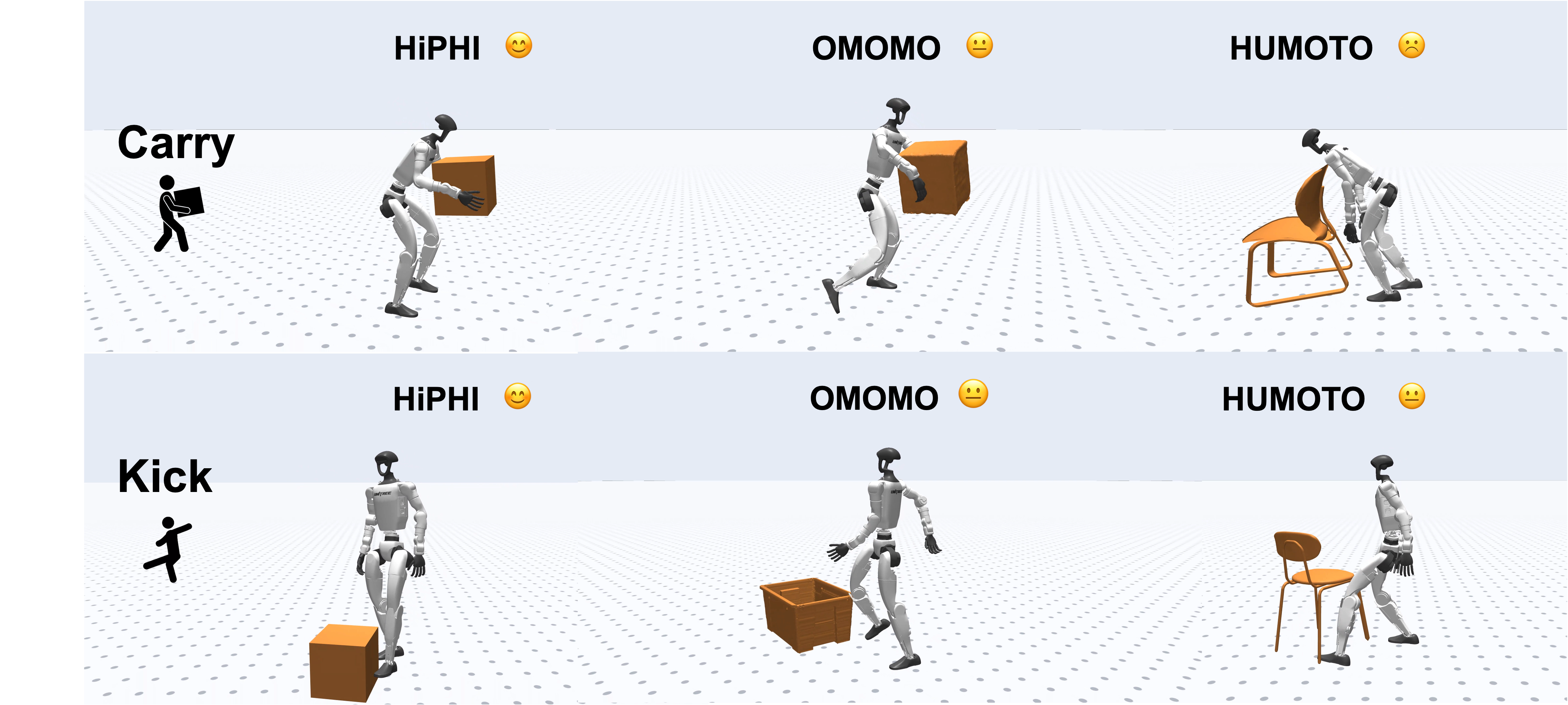}
    \caption{
    \textbf{Qualitative comparisons of motion-with-object tracking across different datasets.}
    The top row shows \textit{carry} interactions, while the bottom row presents \textit{kick} interaction behaviors.
    \ours demonstrates more stable whole-body tracking and better human-object interaction consistency across diverse tasks.
    }
    \label{fig:mwo_qualitative}
\end{figure*}

Figure~\ref{fig:mwo_qualitative} presents qualitative comparisons of motion-with-object tracking across three datasets: \ours, OMOMO, and HUMOTO. 
We visualize two representative interaction categories, including \textit{carry} (first row) and \textit{kick} interactions (bottom row). 
Existing datasets exhibit limited interaction coverage and tracking robustness in certain categories. 
For example, HUMOTO fails to reproduce stable object-aware behaviors in the carry scenario, while OMOMO suffers from reduced interaction fidelity. 
In contrast, \ours produces more stable whole-body motions and more consistent human-object interaction behaviors across diverse tasks.

\end{document}